\pdfoutput=1
\documentclass[11pt]{article}
\usepackage{yandex-research}

\usepackage{amsmath,amsfonts,bm}

\def\eqref#1{equation~\ref{#1}}

\def\1{\bm{1}}

\DeclareMathAlphabet{\mathsfit}{\encodingdefault}{\sfdefault}{m}{sl}
\SetMathAlphabet{\mathsfit}{bold}{\encodingdefault}{\sfdefault}{bx}{n}

\newcommand{\E}{\mathbb{E}}

\DeclareMathOperator*{\argmax}{arg\,max}
\DeclareMathOperator*{\argmin}{arg\,min}

\usepackage{hyperref}
\usepackage{url}
\usepackage{graphicx}
\usepackage{booktabs}
\usepackage{multirow}
\usepackage{microtype}
\usepackage{enumitem}
\usepackage{algorithm}
\usepackage{algpseudocode}
\usepackage{placeins} 
\hypersetup{
  pdftitle={Beyond Foundation Models: PFN Priors Improve Tabular Hyperparameter Optimization},
  pdfauthor={Artem Chistyakov, Renat Sergazinov, Artem Babenko}
}

\newcounter{ablationsec}[subsubsection]
\renewcommand{\theablationsec}{\thesubsubsection.\arabic{ablationsec}}
\newcommand{\ablationsection}[1]{\par\addvspace{0.8\baselineskip}\refstepcounter{ablationsec}%
  \noindent{\bfseries\theablationsec\quad #1}\par\nobreak\addvspace{0.3\baselineskip}}
\newtheorem{assumption}{Assumption}
\newtheorem{proposition}{Proposition}
\newtheorem{lemma}{Lemma}
\providecommand{\qed}{\hfill$\square$}

\title{Beyond Foundation Models: PFN Priors Improve Tabular Hyperparameter Optimization}

\author{%
Artem Chistyakov\textsuperscript{*,1,2},\enspace
Renat Sergazinov\textsuperscript{*,2},\enspace
Artem Babenko\textsuperscript{2,1}%
}
\affiliations{%
\textsuperscript{1}HSE University,\enspace
\textsuperscript{2}Yandex
}

\begin{document}

\maketitle
\begingroup
\renewcommand{\thefootnote}{\fnsymbol{footnote}}
\footnotetext[1]{Equal contribution. Correspondence to \texttt{artemchistyakov03@gmail.com} and \texttt{mc.sergazinov@gmail.com}.}
\endgroup
\addtocontents{toc}{\protect\setcounter{tocdepth}{-1}}

\begin{abstract}
Synthetic priors have so far served one purpose: pretraining tabular foundation models. We ask whether the same priors can improve conventional tabular models, which are trained from scratch on each dataset. We propose PriorTune, a method that meta-tunes hyperparameter configurations on synthetic tasks alone. Offline, for a given model, we fit a pool of configurations on datasets sampled from a published PFN prior and greedily select the subset with the highest winrate over the random baseline. Online, a new dataset fits only the selected configurations, so the gain comes at no extra cost. Empirically, we observe that scaling  both the number of configurations and synthetic datasets has an equal scaling effect on the performance of the final subset of configurations, which we also explain theoretically. On the full BeyondArena benchmark, LightGBM with TabPrep features with our configurations ranks above TabPFN-3 and RealMLP, while being much more efficient at inference. Overall, our results establish existing PFN priors as a scalable source of meta-training data for hyperparameter selection for non-foundational tabular models.

\end{abstract}

\begin{paperlinks}
\paperlinklabel{Code}{https://github.com/yandex-research/PriorTune}
\paperlinklabel{Results}{https://huggingface.co/datasets/mrsergazinov/PriorTune}
\end{paperlinks}

\section{Introduction}
\suppressfloats[t]

Synthetic data drives the development of tabular foundation models. Since TabPFNv1 \citep{hollmann2023tabpfn}, every generation has come with a richer prior: TabPFNv2 added new edge types to the causal graph sampler \citep{hollmann2025tabpfnv2}, TabICLv2 added heavy-tailed random graphs and a filter on the generated datasets \citep{tabiclv2}, and TabPFN-3 scaled the generated datasets to a million rows \citep{tabpfn3}. To date, these priors have mostly served one purpose: pretraining PFNs.

\begin{figure}[t]
    \centering
    \includegraphics[width=\linewidth]{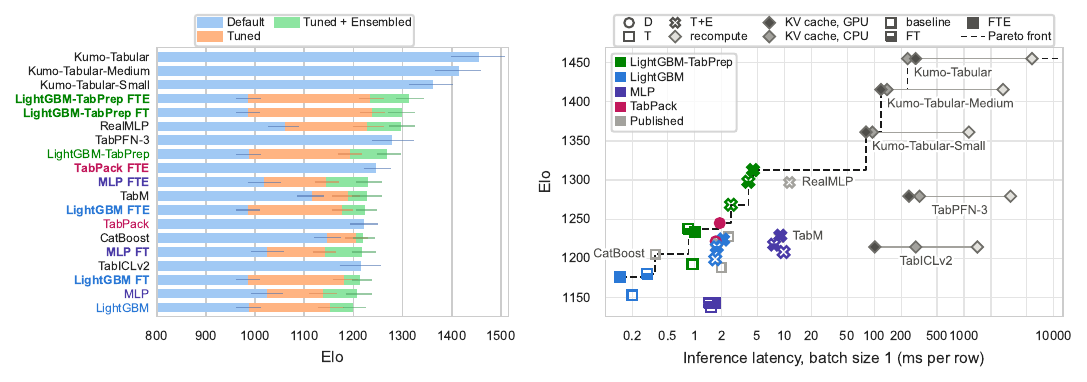}
    \caption{Full BeyondArena. \textbf{Left:} Elo under default, tuned, and tuned-and-ensembled deployment. Our families are coloured; FT (half-colored) and FTE (full-colored) are the subsets meta-tuned under T and T+E. \textbf{Right:} Elo against inference latency at batch size 1. GBDTs run on 16 CPU threads, and DNNs and PFNs on one GPU. For PFNs, we run with recompute (light grey), KV cache on CPU (medium grey), on GPU (dark grey). Protocol in Appendix~\ref{app:latency}.}
    \label{fig:pareto}
\end{figure}

Conventional models, which are trained from scratch on each dataset, have no direct way to use these priors. For such models, e.g. GBDTs and DNNs, hyperparameter selection is the natural way for transferring knowledge across tasks. TabRepo demonstrates that a set of hyperparameter configurations selected on approximately 200 real datasets transfers effectively to unseen datasets \citep{salinas2024tabrepo}. Similarly, RealMLP identifies strong MLP architecture and training defaults through meta-tuning on 118 real datasets \citep{holzmuller2024realmlp}. Both rely on a small, fixed collection of real datasets, which caps the scale and diversity of the meta-training tasks. This raises a natural question: can datasets sampled from a PFN prior surpass real datasets when learning transferable sets of hyperparameter configurations? To test this, we sample 4000 synthetic datasets from TabICLv2 prior, evaluate 4000 hyperparameter configurations on them, and propose a new algorithm to select $K$ best-performing configurations. On a new dataset, we only train the selected subset of $K$ configurations, so the meta-tuning gains transfer for free.

We study four efficient tabular models: LightGBM \citep{ke2017lightgbm}, LightGBM with TabPrep \citep{tschalzev2026tabprep}, an MLP with cosine embeddings \citep{gorishniy2022embeddings}, and TabPack \citep{tabpack}. These four models are near the efficiency-performance Pareto frontier (Figure~\ref{fig:pareto}). We use the TabICLv2 prior, with TabPFNv1 and TabICLv1 in ablations, and evaluate transfer on TabArena's 51 IID datasets \citep{erickson2025tabarena} and BeyondArena's 142 datasets with diverse splitting protocols \citep{beyondarena}.

As a source of meta-training tasks, synthetic priors have three advantages over real datasets. First, supply is unlimited and quality is controllable. Real repositories plateau: on TabRepo's 182 datasets, adding configurations stops helping, whereas on synthetic tasks neither the number of tasks nor the number of configurations has saturated at 4,000 (Section~\ref{sec:ablations}). Second, synthetic priors can be continuously improved: foundational model have shown that improving priors yields better model performance. Third, synthetic tasks rule out test-set contamination by construction. Real data needs an overlap audit: for TabArena, we removed 18 of the 200 TabRepo datasets.

The contributions of this paper are as follows:
\begin{enumerate}[leftmargin=2em]

\item \textbf{Meta-tuning hyperparameters on synthetic data.} We show that existing PFN priors can improve conventional tabular models through offline hyperparameter selection. Per dataset, real tasks help more, but synthetic tasks scale: given equal budget (configurations $\times$ datasets), hyperparameters selected on 4000 synthetic datasets outperform those selected on 182 real TabRepo datasets for every model and every~$K$.

\item \textbf{Improved tuning on real datasets.} On both TabArena and BeyondArena, the selected sets of configurations outperform random sets of the same size. On TabArena, they also outperform random sets of any size, so meta-selection improves both the quality and the efficiency of downstream tuning. On BeyondArena, our LightGBM-TabPrep ranks 7th of 58 methods, ahead of TabPFN-3 and behind only recent foundation models.

\item \textbf{An analysis of scaling and prior choice.} We investigate how the quality of the configuration set depends on the number of synthetic tasks, the candidate pool size, and the generating prior, and derive a transfer bound and a record-count argument that motivate the observed scaling forms. Scaling number of datasets and configurations keeps paying off at 4,000, whereas the choice of prior appears secondary: even the TabPFNv1 prior is competitive with TabICLv2.

\end{enumerate}

\begin{figure}[t]
    \centering
    \includegraphics[width=\linewidth]{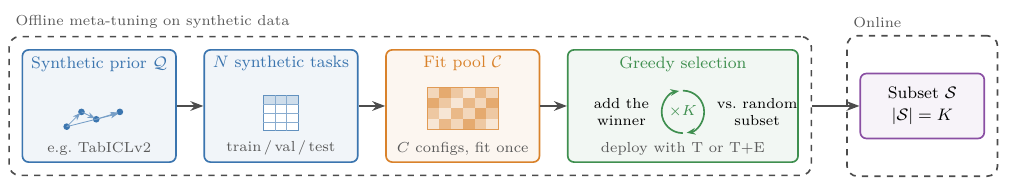}
    \caption{Overview. Offline, we sample hyperparameter configurations from a pool $\mathcal C$, fit them on $N$ synthetic tasks, and greedily select a subset of $K$ by win rate over random subsets under the given deployment rule. Online, we fit the $K$ selected configurations once per task.}
    \label{fig:framework-overview}
\end{figure}

\section{Related Work}

\paragraph{Transfer of hyperparameter configurations.} Good hyperparameter configurations typically transfer across datasets. \citet{feurer2015initializing} warm-start Bayesian optimization with configurations that performed well on similar datasets. Later work drops the similarity and learns one fixed set for all datasets: given a table of scores for every configuration on every dataset, greedily add the configuration that most improves the average best-member score over datasets \citep{wistuba2015sequential,pfisterer2021defaults, winkelmolen2020zeroshot}. TabRepo scales this offline selection to 1,310 configurations on 200 datasets, and AutoGluon 1.0 ships the learned set as its default \citep{salinas2024tabrepo,erickson2020autogluon}. RealMLP meta-learns several model components (e.g. preprocessing, activations) on 118 datasets \citep{holzmuller2024realmlp}. All of this work is bounded by the supply of real data, 25 to 208 datasets. TabRepo reports a saturating effect around 150 datasets or configurations. We keep the greedy forward selection, but run it on synthetic tasks, whose supply is unlimited, and change the scoring function (Section~\ref{subsec:subset-selection}).

A second line transfers knowledge into the optimizer rather than into a set of configurations. One line of work fits the Gaussian-process prior on evaluations from previous tasks \citep{perrone2018scalable,feurer2018transfer,wistuba2021fsbo,wang2024hyperbo,fan2023hyperboplus}. OptFormer pre-trains a text transformer on optimization logs \citep{chen2022optformer}. Recent works such as PFNs4BO and its successors, pretrain PFNs on synthetic optimization tasks \citep{muller2023pfns4bo, yu2026gitbo,rogers2026zeroshot, zhang2026tamo}. All of these search sequentially and need the surrogate model at tuning time. Our configuration set is built once offline and can be evaluated in parallel.

\paragraph{Synthetic priors for tabular data.} Synthetic priors are the engine of tabular foundation models. Among others, the recent TabPFN, TabICL, Mitra, LimiX, Xiaomi-TabLDM, Causilo, and EXAONE Tabular are pretrained on synthetic tables alone \citep{hollmann2023tabpfn,hollmann2025tabpfnv2,tabpfn3,qu2025tabicl,tabiclv2,zhang2025mitra,limix2025,wang2026xiaomitabldm,cho2026causilo,eo2026exaonetabular}. A typical synthetic prior has three parts: a directed graph over variables, an edge function that maps parents to a child, and an observation process. Graphs are random DAGs in TabPFN, Cauchy random graphs with varied connectivity in TabICLv2, and hierarchical causal models in LimiX. Edge functions started as small neural networks and grew to include decision trees and discretization in TabPFNv2, tree ensembles in TabICL, and Gaussian processes, polynomials, and products in TabICLv2  \citep{hollmann2025tabpfnv2,eo2026exaonetabular}. Related in spirit, recent work distills a PFN into a tree ensemble or an MLP  on a given dataset \citep{tanna2026pocket,qin2026gear}. However, distillation requires PFN during the training phase and has yet to demonstrate significant performance improvements.  

\section{Method}
\label{sec:method}

In this section, we present PriorTune, a method for meta-tuning strong hyperparameter configurations of conventional tabular models on synthetic data (Figure~\ref{fig:framework-overview}). Offline, we sample $N$ tasks from a synthetic prior $\mathcal Q$ (Section~\ref{subsec:prior}), fit a pool $\mathcal C$ of $C$ configurations once per task (Section~\ref{subsec:base-model}), and greedily select $K$ of them (Section~\ref{subsec:subset-selection}). Online, a real task fits only the $K$ selected configurations and deploys them by validation, either the best one (\emph{tuned}) or a greedy ensemble (\emph{tuned and ensembled}).

\subsection{Setup}
\label{subsec:setup}
\paragraph{Notation.} We consider regression and classification on tabular data. We denote the synthetic data generator by $\mathcal Q$ and a dataset by $D$. The loss $\ell$ is the native metric of the task, oriented so that lower is better: RMSE for regression, log-loss for multiclass, and $1-$ROC-AUC for binary classification, as in TabArena and BeyondArena. We split each dataset into train, validation, and test subsets $D_{\mathrm{train}}, D_{\mathrm{val}}, D_{\mathrm{test}}$.

\paragraph{Benchmarks.} We use TabArena \citep{erickson2025tabarena} and BeyondArena \citep{beyondarena}, which together cover 155 to 1.3 million rows, 3 to 22,215 features, and i.i.d., grouped, and temporal splits. Table~\ref{tab:bench-data-stats} summarizes both suites. The main results (Section~\ref{sec:experiments}) use the full BeyondArena with every outer train-test split. The ablations of Section~\ref{sec:ablations} use TabArena-Lite, the first train-test split. Each split has up to $8$ inner train-validation splits. No benchmark data enters the selection.

\subsection{Synthetic data generation}
\label{subsec:prior}

For all main experiments, we sample tasks from the TabICLv2 prior \citep{tabiclv2}. Three settings of the prior matter most, and we keep them close to the third pretraining stage of TabICLv2. Dataset size is log-uniform between 400 and 60,000 rows. The feature count is uniform between 1 and a cap that falls from 100 to 20 as the row count grows. The sampled datasets are filtered using both of the available filters: reject if the target node of the graph shares no ancestors with feature nodes, and reject if ExtraTrees cannot beat the constant predictor. We fix the task mix at 2,000 regression, 1,400 binary, and 600 multiclass datasets with 3 to 10 classes, and split each dataset 64/16/20 into train, validation, and test, stratified for classification. Appendix~\ref{app:priors} lists all settings.

\begin{algorithm}[t]
\caption{PriorTune subset selection}
\label{alg:priortune}
\begin{algorithmic}[1]
\Require cached validation and test predictions of every $c\in\mathcal C$ on synthetic tasks $D_1,\dots,D_N$; subset size $K$; number of draws $R$; deployment rule $\mathcal E\in\{\mathrm T,\mathrm{T{+}E}\}$ \eqref{eq:deployment}
\State Draw references $\mathcal R_1,\dots,\mathcal R_R\sim\mathrm{Unif}(\Pi_K)$ and, for every task $i$, bootstrap resamples $D_{\mathrm{val},i}^{(1)},\dots,D_{\mathrm{val},i}^{(R)}$ of its validation rows
\State $\mathcal S_0\gets\emptyset$
\For{$k=1,\dots,K$}
  \For{each candidate $c\in\mathcal C\setminus\mathcal S_{k-1}$}
    \State  \# deploy both with $\mathcal E$ on resample $r$, score on test; $h(a,b)=\1_{a<b}+\tfrac12\1_{a=b}$
    \State $\widetilde U(c)\gets\dfrac1{NR}\displaystyle\sum_{i=1}^N\sum_{r=1}^R h\big(L_i^{(r)}(\mathcal S_{k-1}\cup\{c\}),\,L_i^{(r)}(\mathcal R_r)\big)$
  \EndFor
  \State $c_k\gets\argmax_c\widetilde U(c)$, ties broken by the lowest configuration id
  \State $\mathcal S_k\gets\mathcal S_{k-1}\cup\{c_k\}$
\EndFor
\State \Return $\mathcal S_K$
\end{algorithmic}
\end{algorithm}

\subsection{Models}
\label{subsec:base-model}

For our main runs, we use LightGBM \citep{ke2017lightgbm}, LightGBM with TabPrep \citep{tschalzev2026tabprep}, MLP with cosine embeddings \citep{gorishniy2022embeddings}, and TabPack \citep{tabpack}. We select these models for their tradeoff between training and inference speed and performance (Figure~\ref{fig:pareto}). Each configuration in the pool $\mathcal C$ with $C=4{,}000$ is an independent draw from the model's published search space, which we list in Appendix~\ref{app:comp-hps}. TabPack is a packed ensemble of MLPs, so its unit of sampling is a single MLP configuration: it shares the pool of $C=4{,}000$ MLP configurations, from which we select its $K=64$ members.

\subsection{Subset selection}
\label{subsec:subset-selection}

The selection mimics the deployment rule and scores against random subsets of the same size. Following TabArena and BeyondArena, we distinguish two deployment rules: \emph{tuned} (T), where the validation-best member is deployed, and \emph{tuned and ensembled} (T+E), where validation builds a greedy ensemble \citep{caruana2004ensemble}. We implement the whole procedure as a greedy forward selection algorithm: at each iteration starting from an empty set, we add a configuration, score it under the given deployment rule, and select the one that yields the highest winrate against the random subset (Algorithm~\ref{alg:priortune}). Appendix~\ref{app:theory} states the objective formally. Below we motivate each choice on a small example (see Section~\ref{subsec:selection-ablation} for ablation).

\begin{table}[b]
\caption{A toy example under the \emph{tuned} rule. Four configurations, three synthetic datasets $D_1$--$D_3$ used for selection ($D_1$ scored by RMSE, the rest by log-loss) and one held-out dataset $D_4$ (log-loss). Cells are validation / test losses (lower is better); the last row is the test loss of the deployed member of a random reference subset. We select a subset of $K=2$ from the empty set.}
\label{tab:toy}
\begin{center}
\small
\begin{tabular}{l cc cc cc c cc}
\toprule
& \multicolumn{2}{c}{$D_1$} & \multicolumn{2}{c}{$D_2$} & \multicolumn{2}{c}{$D_3$} & & \multicolumn{2}{c}{$D_4$, held out} \\
\cmidrule(lr){2-3}\cmidrule(lr){4-5}\cmidrule(lr){6-7}\cmidrule(lr){9-10}
Configuration & val & test & val & test & val & test & & val & test \\
\midrule
A & 105 & 103 & 0.28 & 0.36 & 0.60 & 0.45 & & 0.38 & 0.47 \\
B & 96 & 96 & 0.30 & 0.31 & 0.53 & 0.53 & & 0.41 & 0.41 \\
C & 95 & 97 & 0.34 & 0.34 & 0.55 & 0.55 & & 0.40 & 0.40 \\
D & 60 & 60 & 0.29 & 0.40 & 0.52 & 0.60 & & 0.39 & 0.50 \\
\midrule
Random reference & -- & 100 & -- & 0.32 & -- & 0.50 & & -- & 0.42 \\
\bottomrule
\end{tabular}
\end{center}
\end{table}

Table~\ref{tab:toy} gives a toy instance under the tuned rule: four configurations A to D, three synthetic tasks $D_1$ to $D_3$ for selection, one held-out task $D_4$ standing for a real dataset, and a subset size of $K=2$.

\begin{figure}[t]
    \centering
    \includegraphics[width=\linewidth]{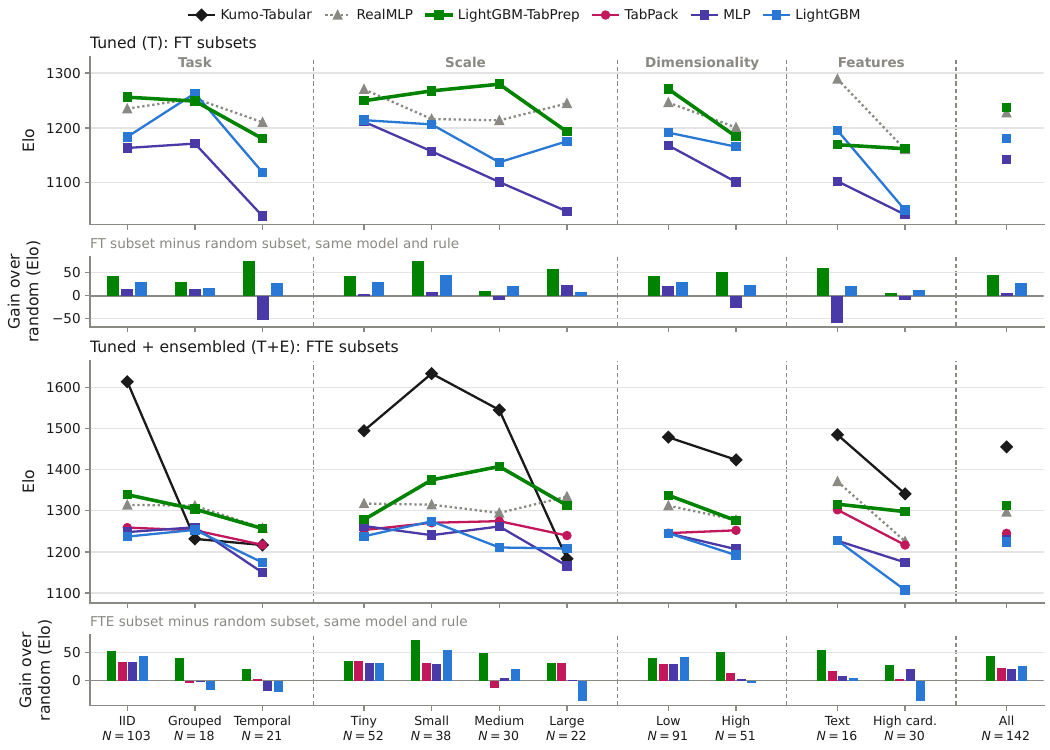}
    \caption{Elo on the full BeyondArena by dataset group. \textbf{Top:} tuned (T), our models with their FT subsets. \textbf{Bottom:} tuned and ensembled (T+E), our models with their FTE subsets. Under each panel, the bars show the Elo gain of each meta-tuned subset over the same model with its random subset.}
    \label{fig:main}
\end{figure}

\paragraph{Selecting on validation, scoring on test.} Under both deployment rules, the model (or ensemble) is selected on the validation split and scored on the test split. Our objective mimics this, which selects configurations that are stable across validation and test. The alternative, selecting on test directly or on a single out-of-fold score as TabRepo and AutoGluon do \citep{salinas2024tabrepo}, hides that instability. In Table~\ref{tab:toy}, A and D are unstable: their validation and test losses disagree on most datasets, whereas B and C are stable. Both selections pick B first, then diverge: test-only selection returns $\{\mathrm B,\mathrm A\}$, validation-then-test returns the stable $\{\mathrm B,\mathrm C\}$. On the held-out $D_4$, $\{\mathrm B,\mathrm A\}$ deploys A and loses; $\{\mathrm B,\mathrm C\}$ deploys C and wins.

\paragraph{Comparing against the random baseline.} The alternative is to use losses directly: TabRepo adds the configuration that most lowers the mean error of the best member over datasets \citep{salinas2024tabrepo}. Raw losses mix metrics and scales across tasks. In Table~\ref{tab:toy}, any rule on raw losses selects D first, since it cuts the RMSE on $D_1$ by 36 points, which outweighs every other gain in the table, but D has the worst test loss on every other dataset. Ranking objectives are preferred over regression objectives for the same reason in offline model-based optimization \citep{tan2025ranking}.

\paragraph{Stability.} The third design choice is to resample both the validation split and the random reference. A single validation split decides which member is deployed: in Table~\ref{tab:toy}, C beats B on the validation split of $D_1$ by one RMSE point, and another validation sample may deploy B instead. We therefore draw $R=16$ bootstrap resamples of the validation rows, paired with $R$ independent random references $\mathcal R_r$; with cached predictions the cost is negligible, and averaging over references is unbiased with variance $O(1/R)$ (Proposition~\ref{prop:reference}).

\section{Main results}
\label{sec:experiments}

\paragraph{Setup.} We evaluate on the full BeyondArena core (142 datasets) under the benchmark's protocol: every outer train-test split, 8-fold bagging within each, native metrics, and $K=25$ configurations per dataset (the official budget). TabPack fits its $K=64$ members. FT and FTE denote the subsets meta-tuned for T and T+E. A random subset has 25 draws from the same search space. We rank methods by Elo, which counts pairwise wins and losses against all other methods. Figure~\ref{fig:main} reports Elo within each BeyondArena dataset group. Appendix~\ref{app:main-improvement} reports alternative metric, the median improvement over the tuned MLP.

\paragraph{State of the art and gains over random subsets.} Under T, LightGBM-TabPrep with its FT subset has the highest Elo of all tuned methods (Figure~\ref{fig:pareto}). Under T+E, LightGBM-TabPrep with its FTE subset ranks 7th, above RealMLP and TabPFN-3. The 6 methods above it are pretrained foundation models: TabPFN-3.5 (fast and regular), Causilo, and 3 sizes of Kumo-Tabular \citep{nvidia2026kumo} (Figure~\ref{fig:elo-full} in the appendix shows the full leaderboard). Across groups (Figure~\ref{fig:main}), LightGBM-TabPrep leads RealMLP on small and medium datasets, and RealMLP leads on tiny and large datasets and on text features. Compared to Kumo-Tabular, LightGBM-TabPrep loses on i.i.d.\ splits and on tiny to medium datasets, but leads it on grouped and temporal splits and on large datasets (1313 against 1183 Elo). The bars of Figure~\ref{fig:main} show the gain of each meta-tuned subset over its random subset. LightGBM-TabPrep gains in every group, by 21 to 72 Elo under T+E and by 5 to 76 under T. The other models gain 5 to 28 Elo overall, but lose to their random subsets in some groups, most often on grouped and temporal splits. These gains cost nothing at deployment, and Section~\ref{subsec:ladder} shows that, on TabArena, random subsets of any size up to the full pool of $4{,}000$ do not close the gap.

\paragraph{Accuracy and inference cost.} The right panel of Figure~\ref{fig:pareto} plots Elo against the latency of a single-row prediction. Each PFN appears in three modes: recompute, and a key-value (KV) cache kept on the CPU or on the GPU. Up to 1313 Elo, the Pareto front consists of LightGBM, CatBoost, TabPack, and LightGBM-TabPrep. LightGBM-TabPrep with its FTE subset under T+E reaches 1313 Elo at 4.4\,ms. The references with a similar Elo are 2.6 to 740 times slower: RealMLP under T+E reaches 1297 Elo at 11\,ms, TabPFN-3 1279 at 243 to 3{,}308\,ms, and TabICLv2 1215 at 102 to 1{,}412\,ms. Above 1313 Elo, the front consists of the 3 sizes of Kumo-Tabular with a KV cache, from 1361 Elo at 81\,ms to 1455 Elo at 236\,ms. Meta-tuning changes only which configurations are fitted, so the latency stays within a factor of 1.8 of the random subset.

\section{Ablations}
\label{sec:ablations}

\begin{figure}[t]
    \centering
    \includegraphics[width=\linewidth]{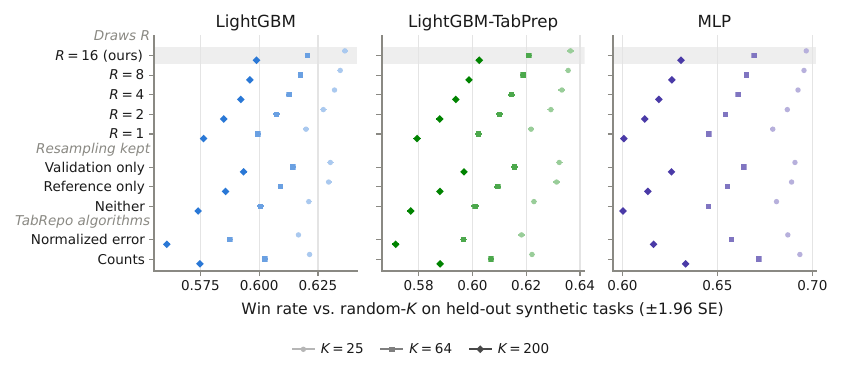}
    \caption{Selection rules, synthetic out-of-fold, T deployment. Each of 200 repetitions selects on 3,000 random synthetic tasks and deploys on the other 1,000. Points are win rates against random-$K$ baselines that no selection saw, $\pm1.96$ standard errors over repetitions; the shaded row is our rule. Values in Table~\ref{tab:selection}.}
    \label{fig:selection}
\end{figure}

All ablations use subsets selected under T and report the win rate of the meta-tuned subset against random subsets of the same size on TabArena (details in Appendix~\ref{app:setup-ablations}). We use T because T+E selection runs a Caruana ensemble per candidate at every step, which is expensive. TabPack is absent: it is an online ensemble of MLP members, so the MLP configurations cover it. The main text shows $K=200$, TabArena's default tuning budget, and we include $K=25, 64$ in Appendix~\ref{app:ablation-grids}.

\subsection{Selection rule}
\label{subsec:selection-ablation}

Figure~\ref{fig:selection} ablates the choices of Section~\ref{subsec:subset-selection}. To avoid potential data leak, we evaluate out-of-fold on synthetic tasks: randomly select 3,000 datasets, deploy under T on the other 1,000, repeat 200 times. All rules choose from the same 4,000 configurations and are scored against the same random-$K$ baselines (Appendix~\ref{app:setup-selection}).

\begin{figure}[t]
    \centering
    \includegraphics[width=\linewidth]{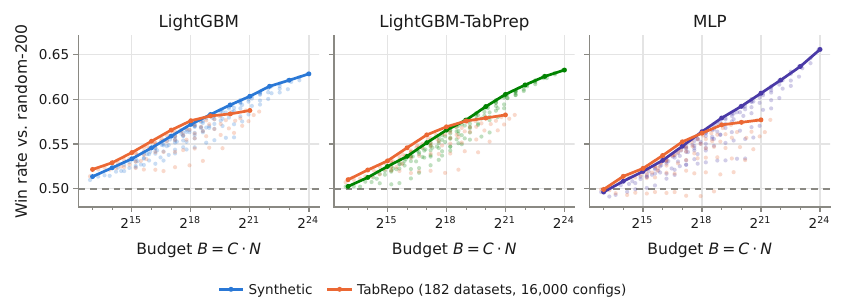}
    \caption{TabArena, T deployment, win rate against a random subset of the same size. Synthetic versus TabRepo meta-training data at $K=200$. Dots are $(C,N)$ cells; lines are the best cell within each factor-of-two budget bin. For TabRepo (182 datasets), we evaluated 16,000 configurations, so its curve stops at $B=16{,}000\times182$.}
    \label{fig:budget}
\end{figure}

\paragraph{Comparing to TabRepo.} The TabRepo algorithm scores the best member of a subset directly, with no validation-then-test step, so we give it one score per task on the joined validation and test rows. We replace the raw loss with TabRepo's normalized error, $(\ell-\ell_{\mathrm{best}})/(\ell_{\mathrm{median}}-\ell_{\mathrm{best}})$ clipped to $[0,1]$ per task \citep{salinas2024tabrepo}, which removes the metric scale. From the empty set, each step adds the configuration that minimizes the mean over tasks of the best normalized error in the subset. It trails $R=16$ for every model and $K$, by 0.010 to 0.038: even with the scale gone, the loss shape steers the selection into suboptimal regions. A naive improvement is to count instead the tasks on which a candidate improves the best member (i.e., a win rate against the current subset). The improved version beats the base in all nine cases, by 0.004 to 0.017. However, it still trails $R=16$ by 0.014 to 0.024 for LightGBM and LightGBM-TabPrep, where scoring the best member hides configurations whose validation rank does not hold on test, and ties it on the MLP ($-0.003$ to $+0.003$).

\paragraph{Resampling validation and references.} The plain hill-climb, with one validation split and one reference, trails $R=16$ by 0.014 to 0.031. If we resample (16 times) just the validation, we close 60 to 84\% of the gap. If we resample just the random references, we close 41 to 62\%. Both matter, but validation noise appears to matter more. The win rate rises monotonically in $R$ for every model and $K$: by 0.015 to 0.030 from $R=1$ to $R=16$, most at $K=200$.

\subsection{Synthetic versus real meta-training data}
\label{subsec:tabrepo}

TabRepo \citep{salinas2024tabrepo} is the standard real-data source for meta-tuning: 200 datasets, of which we drop 18 that overlap with or share a source with TabArena (listed in Appendix~\ref{app:ablation-grids}). We fit 16,000 configurations (our pool of 4,000 and 12,000 further draws) on the 182 datasets and select on them with the same protocol. Figure~\ref{fig:budget} compares the two sources at equal budget $B=C\cdot N$. Per task, real data is worth more at smaller budgets: up to $2^{18}$ subsets selected on the synthetic data show lower win rates, but from $2^{19}$ to $2^{20}$ on, synthetic data overtakes TabRepo. TabRepo saturates: over its last two doublings, to 16,000 configurations, its curve gains at most 0.008. We attribute this to the TabRepo size: with 182 datasets, the subsets start to overfit the source datasets. Synthetic data allows us to sidestep this, because we can continue scaling the data size indefinitely: at $2^{24}$ it ends 0.02 to 0.08 above TabRepo's best for every model and every $K$. At $K=25$ and $64$ the crossover lies between $2^{20}$ and $2^{22}$ (Appendix~\ref{app:ablation-grids}).

\subsection{Meta-tuning versus more random search}
\label{subsec:ladder}

\begin{figure}[t]
    \centering
    \includegraphics[width=\linewidth]{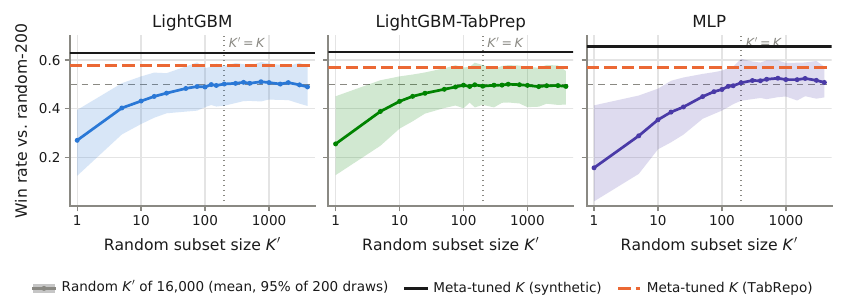}
    \caption{TabArena, T deployment, $K=200$. Random subsets of growing size $K'$ against the random-$K$ baseline. Solid curve and band: mean and 95\% CI over 200 random subsets of size $K'$. Horizontal lines are the $K$ configurations meta-tuned on all 4,000 configurations, with 4,000 synthetic tasks or the 182 TabRepo datasets, averaged against the same baselines.}
    \label{fig:ladder}
\end{figure}

A cheaper alternative to meta-tuning is to fit more random configurations. Figure~\ref{fig:ladder} scores random subsets of size $K'$ against the random-$K$ baseline (details in Appendix~\ref{app:setup-random}). The mean curve rises, peaks at 0.50 to 0.53 for $K'$ between 500 and 750, and falls to 0.49 to 0.51 at $K'=4{,}000$: larger random subsets overfit the validation split. At $K=25$ and $64$, the mean curve peaks higher, at 0.51 to 0.60, but the meta-tuned subset still sits at least 0.07 above it (Appendix~\ref{app:ablation-grids}). For a practitioner with a fixed budget, the 200 meta-tuned configurations replace a random search over 4,000 configurations that fits twenty times more models and still scores lower on average.

\subsection{Choice of prior}
\label{subsec:prior-ablation}

\begin{figure}[t]
    \centering
    \includegraphics[width=\linewidth]{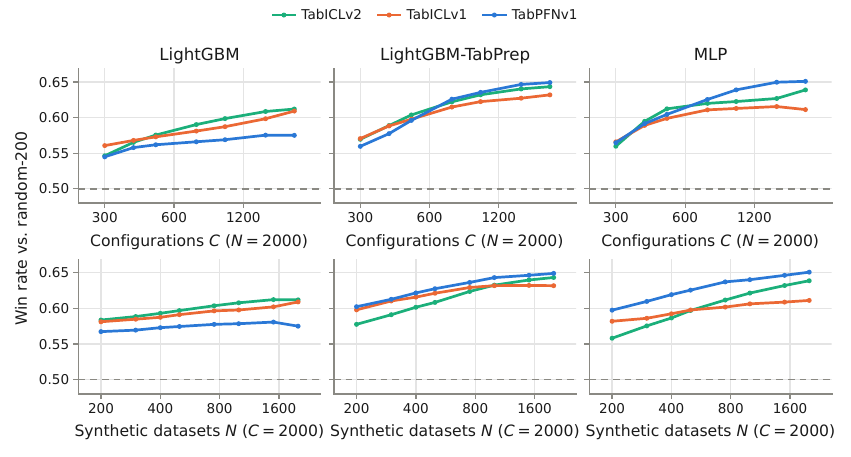}
    \caption{TabArena, T deployment, win rate against a random subset of the same size, $K=200$. Three priors, each with 2,000 configurations on 2,000 synthetic tasks, for LightGBM, LightGBM-TabPrep, and the MLP. Top: as $C$ grows at $N=2{,}000$; bottom: as $N$ grows at $C=2{,}000$. Win rates are averaged over 2,000 repetitions.}
    \label{fig:priors}
\end{figure}

\begin{figure}[t]
    \centering
    \includegraphics[width=\linewidth]{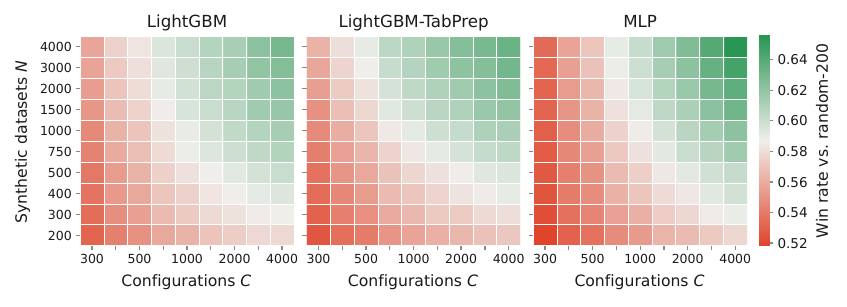}
    \caption{TabArena, T deployment, win rate against a random subset of the same size over the $(C,N)$ grid, $K=200$. Win rates are averaged over 2,000 repetitions.}
    \label{fig:scaling}
\end{figure}

We repeat the selection on synthetic tasks from the TabICLv1 \citep{qu2025tabicl} and TabPFNv1 \citep{hollmann2023tabpfn} priors (Appendix~\ref{app:priors}), with $2{,}000$ configurations on $2{,}000$ datasets for LightGBM, LightGBM-TabPrep, and the MLP (Figure~\ref{fig:priors}). With all 2,000 configurations and tasks, the prior matters most for plain LightGBM: TabICLv2 leads at every $K$, by up to 0.09 over TabPFNv1 (0.644 against 0.551 at $K=25$). With TabPrep, the three priors are within 0.02 of each other at every $K$. For the MLP, TabICLv2 wins at $K=25$ and $64$ and TabPFNv1 at $K=200$. No prior wins everywhere, and the prior of TabPFNv1 appears to be competitive. We use TabICLv2, the newest prior, for the main experiments.

\subsection{Scaling}
\label{subsec:scaling}

Figure~\ref{fig:scaling} shows the win rate over the full $(C,N)$ grid. It rises at almost every step along both axes, and neither axis has saturated at $4{,}000$. Appendix~\ref{app:theory} motivates two forms: the estimation error over $N$ synthetic tasks decays as $1/\sqrt N$ (Proposition~\ref{prop:transfer}), and the number of improvements available in a pool grows as $\log C$ (Proposition~\ref{prop:records}). We fit both forms to the curves at $C=4{,}000$ and $N=4{,}000$ and obtain $R^2$ between 0.92 and 1.00 for $a-b/\sqrt N$ and between 0.90 and 0.98 for $a+b\log_2C$, over all three models and all $K$. In Figures~\ref{fig:scaling-grid-datasets} and~\ref{fig:scaling-grid-configs} in the appendix, the curve in $C$ is steeper than the curve in $N$ for every model: at $K=200$, a doubling of $C$ is worth 1.2 to 1.7 doublings of $N$, and the curve in $N$ tends to flatten above about 1,000 tasks. Once there are a few thousand synthetic tasks, more configurations are the better use of a fixed budget. Tasks reduce the estimation error of the objective, whereas configurations add candidates worth finding.

\subsection{Selected hyperparameters}
\label{subsec:hp}

\begin{figure}[t]
    \centering
    \includegraphics[width=\linewidth]{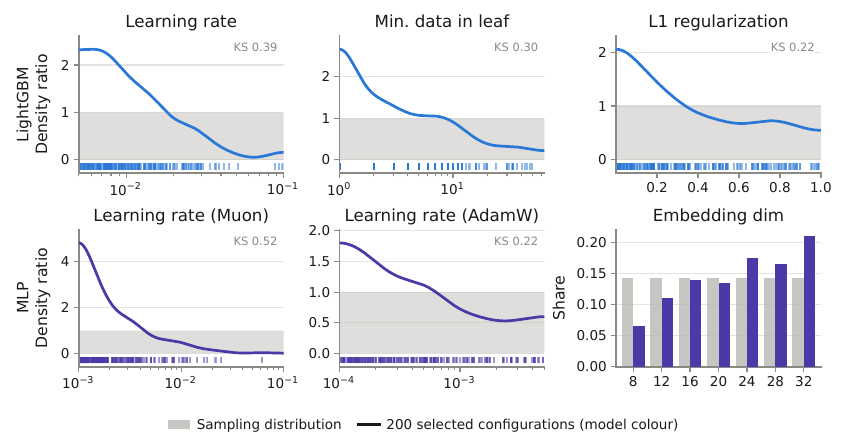}
    \caption{Selected hyperparameters against their sampling distribution: $K=200$, T regime, selected from the full pool of 4000 configurations on 4000 synthetic datasets. Lines: density of the 200 selected configurations divided by the sampling density (grey, flat at 1); ticks: selected values; bars: shares (grey: sampling). All hyperparameters, also of LightGBM-TabPrep, are in Figures~\ref{fig:hp-lgbm-pure}--\ref{fig:hp-mlp}.}
    \label{fig:hp}
\end{figure}

Figure~\ref{fig:hp} compares the selected configurations with the distribution that we sample them from. For LightGBM, the selected configurations prefer more expressive trees, with a small minimum data in a leaf and weak L1 regularization, together with a small learning rate: 21\% of the selected learning rates are below the 0.1 quantile of the sampling distribution. For the MLP, the pattern is similar. The selected configurations prefer small learning rates, with 45\% of the Muon learning rates below the 0.1 quantile, and larger embeddings: the largest dimension, 32, has a share of 21\% against 14\% in the sampling. For these hyperparameters, the range or the shape of the search space is thus possibly a poor fit, and a wider or reshaped space can contain better configurations.

\section{Conclusion}

We have shown that the synthetic priors built for PFN pretraining are a scalable source of meta-training tasks for conventional tabular models. PriorTune selects a subset of hyperparameter configurations on synthetic tasks alone, by win rate against random subsets under the deployment rule.  We show that when scaled the synthetic tasks beat real datasets of TabRepo at equal budgets. The selected subsets beat random subsets of any size, improve LightGBM, LightGBM-TabPrep, the MLP, and TabPack on the TabArena and BeyondArena benchmarks, and make LightGBM-TabPrep the highest-ranked model in Elo on BeyondArena that is not a pretrained foundation model, above TabPFN-3 and RealMLP, at lower latency than both. The released subsets are drop-in search spaces: a practitioner replaces random search over the published space with the $K$ selected configurations.

\clearpage
\section{Limitations}

We do not apply PriorTune to tabular foundation models. They are pretrained on tasks from the same kind of prior, so the selection can prefer configurations that change them the least. Newer priors do not always help PriorTune: the newest prior helps plain LightGBM most but gives no clear gain for LightGBM-TabPrep and the MLP (Section~\ref{subsec:prior-ablation}). The grid is fitted once per model and search space, so a new search space needs a new grid. The synthetic tasks have at most 60,000 rows and 100 features, so the selection does not see tasks as large as the largest real datasets. The ablations use the tuned rule only, because selection for ensembles is expensive. The error bars measure variation across repetitions, not uncertainty over new datasets. Our theory motivates the forms of the scaling curves but does not prove them; the fits are empirical.

\subsection*{AI use statement}

We used generative AI tools (Claude Code, Codex) to implement parts of the experiment and analysis code, to draft and edit parts of the text, and to search for and format references. We did not use them to generate data, develop the theory, prove claims, propose hypotheses, or design the method and experiments. The authors checked and tested all code, cross-checked every number against the experiment logs, verified all references, and reviewed all text. We take full responsibility for the content of this work.

\subsection*{Reproducibility statement}

Section~\ref{sec:method} and Algorithm~\ref{alg:priortune} specify the selection procedure and its fixed settings ($R=16$, tie rule, 40 Caruana steps). Appendix~\ref{app:comp-setup} gives the evaluation protocol on both benchmarks, the repetition scheme of the ablations, the compute used, and the latency measurement. Appendix~\ref{app:comp-hps} lists the complete hyperparameter search space of every model. Appendix~\ref{app:priors} describes the three synthetic priors and Table~\ref{tab:prior-settings} the exact sampling settings and seeds. Appendix~\ref{app:ablation-grids} lists the TabRepo datasets excluded for overlap with TabArena. Appendix~\ref{app:theory} states all assumptions and gives complete proofs of the propositions. The benchmarks (TabArena, BeyondArena, TabRepo) are public. The supplementary material contains the code, the selected subsets for every model and $K$, the seeds, and the score caches of all fits, from which every table and figure can be regenerated.

\clearpage
\bibliography{iclr2027_conference}
\bibliographystyle{unsrtnat}

\newpage
\appendix
\renewcommand{\contentsname}{Appendix contents}
\addtocontents{toc}{\protect\setcounter{tocdepth}{3}}
\tableofcontents
\newpage

\section{Computation}
\label{app:comp}

\subsection{Setup}
\label{app:comp-setup}

\subsubsection{Main results}
\label{app:setup-main}

\paragraph{Datasets.} Table~\ref{tab:bench-data-stats} shows distribution of features and sample sizes of both benchmarks. For full list of datasets, their selection criteria, and description, see the original work \citep{erickson2025tabarena, beyondarena}.

\begin{table}[h]
\caption{Benchmark statistics. Row and feature counts are per dataset (minimum / median / maximum).}
\label{tab:bench-data-stats}
\begin{center}
\small
\begin{tabular}{lcc}
    \toprule
     & TabArena & BeyondArena \\
    \midrule
    Datasets & 51 & 142 \\
    Binary / multiclass / regression & 30 / 8 / 13 & 73 / 25 / 44 \\
    Splits: i.i.d.\ / grouped / temporal & 51 / 0 / 0 & 103 / 18 / 21 \\
    Rows & 748 / 6,497 / 150,000 & 155 / 5,000 / 1,296,316 \\
    Features & 4 / 17 / 1,776 & 3 / 23 / 22,215 \\
    \bottomrule
\end{tabular}
\end{center}
\end{table}

\paragraph{Evaluation.} On BeyondArena, we use every outer split of the core benchmark. Within each outer split, every method is bagged over the 8 inner folds, and each configuration is trained once per fold. T deploys the validation-best configuration; T+E fits a 40-step Caruana ensemble on the same validation predictions. Both score the held-out test split with the native metric.

\paragraph{Subsets.} FT subsets are selected with the T objective and FTE subsets with the T+E objective (40 Caruana steps inside the selection), both with $R=16$ on all 4,000 configurations and 4,000 synthetic tasks. No benchmark data enters the selection. TabPack's FTE subset is the $K=64$ subset selected with the T+E objective on the MLP pool and deployed as one pack of those members.

\subsubsection{Ablations}
\label{app:setup-ablations}

All ablations select under the T deployment rule, deploy under T, and report the win rate against a random-$K$ baseline.

\ablationsection{Selection rule (Section~\ref{subsec:selection-ablation})}
\label{app:setup-selection}

\emph{Protocol.} We run 200 repetitions. Each splits the 4,000 synthetic tasks at random into 3,000 selection tasks and 1,000 held-out tasks. Every rule selects $K$ out of 4,000 configurations on the selection tasks and is deployed under T on the held-out tasks. There it is compared with 200 random-$K$ baselines, and its win rate is averaged over them. All rules in a repetition use the same tasks and baselines. Error bars in Figure~\ref{fig:selection} are $\pm1.96$ standard errors of the win rate over the 200 repetitions.

\emph{Our rule.} It draws, once per repetition, 16 bootstrap resamples of each task's validation rows and 16 random-$K$ references $\mathcal R_r$. $R<16$ pairs the first $R$ resamples with the first $R$ references. \emph{Validation only} pairs the 16 resamples with $\mathcal R_1$, and \emph{reference only} pairs the original validation split with the 16 references.

\emph{TabRepo algorithms.} They use no validation resamples and no references. They score the best member of a subset on each dataset's joined validation and test rows, with the native metric recomputed on the union. The first version adds, at each step, the configuration that minimizes the mean over datasets of the best normalized error, $(\ell-\ell_{\mathrm{best}})/(\ell_{\mathrm{median}}-\ell_{\mathrm{best}})$ clipped to $[0,1]$, with $\ell_{\mathrm{best}}$ and $\ell_{\mathrm{median}}$ taken over the 4,000 configurations. The second starts from the configuration with the lowest mean rank, then adds the one that improves the best member on the most tasks. Both break ties by a random configuration order drawn per repetition.

\ablationsection{Synthetic versus real meta-training data (Section~\ref{subsec:tabrepo})}
\label{app:setup-tabrepo}

\emph{Cells.} A cell $(C,N)$ selects $K$ of $C$ configurations on $N$ meta-training tasks by Algorithm~\ref{alg:priortune} with $R=16$. Synthetic cells use our pool of 4,000 configurations and 4,000 synthetic tasks. TabRepo cells use the 182 TabRepo datasets and a bank of 16,000 configurations: our pool of 4,000 plus 12,000 further draws from the same search spaces, fitted on TabRepo and TabArena. Validation resamples are drawn once per task and shared across repetitions. Ties in validation selection and in the greedy step go to the lowest configuration id.

\emph{Repetitions.} We run $2{,}000$ repetitions, shared by all cells. A repetition permutes the configurations and the tasks once; each cell takes the first $C$ configurations and $N$ tasks of that permutation.

\emph{Evaluation.} The selected subset is deployed on TabArena under T and compared with the first $K$ configurations of the repetition's permutation of our 4,000-configuration pool. Both synthetic and TabRepo cells share this baseline. Also both synthetic and TabRepo $C=K$ cells therefore have $0.5$ by construction.

\emph{Reported values.} Each win rate is the mean over the $2{,}000$ repetitions, its standard error is below 0.003 in every cell and is not drawn. It measures variation across repetitions on the fixed benchmark, not uncertainty over new datasets. In Figure~\ref{fig:budget}, each cell $(C,N)$ is a dot at its budget $B=C\cdot N$, so the dots above one $B$ are all cells with that budget. The line joins the best cells within each $B$.

\ablationsection{Meta-tuning versus more random search (Section~\ref{subsec:ladder})}
\label{app:setup-random}

\emph{Meta-tuned lines.} The synthetic line is the cell $C=N=4{,}000$ of Appendix~\ref{app:setup-tabrepo}. The TabRepo line selects on all 182 TabRepo datasets from our pool of 4,000 configurations, with the same protocol, repetitions, and baselines. Each line is the mean win rate over the $2{,}000$ repetitions.

\emph{Random subsets.} For every $K'$, we draw 200 subsets of $K'$ configurations uniformly from the 16,000-configuration bank, independently across $K'$, and deploy each under T on TabArena. Each subset is compared with all $2{,}000$ random-$K$ baselines of the meta-tuned lines, and its win rate is averaged over them. The band therefore reflects only the draw of the random subset: we report the mean over the 200 subsets and the 95\% CI band.

\ablationsection{Choice of prior (Section~\ref{subsec:prior-ablation})}
\label{app:setup-priors}

\emph{Cells.} For LightGBM, LightGBM-TabPrep, and the MLP, a pool of 2,000 configurations is fitted on 2,000 tasks (datasets) from each of TabICLv2, TabICLv1, and TabPFNv1 (Appendix~\ref{app:priors}). Within a model, the configuration pool and the TabArena evaluation are shared, so only the priors affect the selection of the $K$ configurations. A cell $(C,N)$ selects $K$ of $C$ configurations on $N$ tasks by Algorithm~\ref{alg:priortune} with $R=16$.

\emph{Repetitions and evaluation.} As in Appendix~\ref{app:setup-tabrepo}: $2{,}000$ repetitions shared by all cells, each permuting the configurations and the tasks once. The selected subset is deployed on TabArena under T and compared with the first $K$ configurations of the permutation. Win rates are means over the repetitions, with standard errors below 0.003.

\ablationsection{Scaling (Section~\ref{subsec:scaling})}
\label{app:setup-scaling}

\emph{Cells.} The synthetic cells of Appendix~\ref{app:setup-tabrepo}, with all 4,000 configurations and 4,000 synthetic tasks available: $C$ and $N$ range from $K$ to 4,000, with the same $2{,}000$ repetitions, baselines, and reported values.

\emph{Fits.} We fit $a-b/\sqrt N$ to the win rate against $N$ at $C=4{,}000$ and $a+b\log_2 C$ to the win rate against $C$ at $N=4{,}000$, by least squares on the means; the $C=K$ cell, $0.5$ by construction, is left out of the second fit.

\ablationsection{Selected hyperparameters (Section~\ref{subsec:hp})}
\label{app:setup-hp}

\emph{Subset.} We take the $K=200$ subset selected under T on all 4,000 configurations and 4,000 synthetic tasks. For TabPrep settings, only the selected configurations with TabPrep enabled count.

\emph{Distributions.} For each hyperparameter, the sampling distribution $F$ is given in Appendix~\ref{app:comp-hps}. For a continuous hyperparameter, we plot the density of the selected values divided by the sampling density, $f_{\mathrm{sel}}(x)/f(x)$. The sampling distribution is then flat at 1, and a value above 1 is selected more often than it is sampled from the prior. We estimate this ratio with a Gaussian kernel (Scott bandwidth) and plot it as the coloured line. The Kolmogorov--Smirnov (KS) distance is $\sup_x|\hat F_{\mathrm{sel}}(x)-F(x)|$, where $\hat F_{\mathrm{sel}}$ is the empirical distribution function of the selected values. It is in $[0,1]$: 0 when the selected values follow the sampling distribution exactly, and larger when they move away from it. Categorical hyperparameters are shown as shares.

\subsubsection{Compute}
\label{app:compute}

We report compute in A100 hours. CPU work is converted at AWS on-demand prices:\footnote{For GPU, p4d.24xlarge instance, \$21.96 per hour for eight A100s. For CPU, c7i.16xlarge instance, \$2.86 per hour for 64 vCPUs. \url{https://aws.amazon.com/ec2/pricing/on-demand/}, US East (N.~Virginia), accessed October 2026.} 1 A100 hour costs as much as 61.5 vCPU-hours. Table~\ref{tab:compute} summarizes the costs.

\paragraph{PriorTune, once per model.} \emph{(i) Configuration$\times$task grid.} We fit MLPs with the tensorized code of TabPack \citep{tabpack}: one fit takes 0.1 A100 seconds on average on the synthetic tasks, so the 16 million fits of the $4{,}000\times4{,}000$ grid take 400 A100 hours. One LightGBM-TabPrep fit takes about 5 core-seconds, so its grid takes 20,000 core-hours, about 300 A100 hours. \emph{(ii) Selection.} Under T, selecting a subset from the cached predictions takes seconds. Under T+E, selecting 200 configurations takes about 10 A100 hours.

\paragraph{PriorTune, per dataset.} A new dataset fits only the $K$ selected configurations, so the cost equals that of tuning with $K$ random ones. As a proxy, we take the benchmark's tuned LightGBM on BeyondArena-Lite (26 configurations, 8 folds, 16 vCPUs): 8 vCPU-hours per dataset on average (median 2.4), that is, 0.1 A100 hours (median 0.04).

\paragraph{PFNs, once.} Pretraining TabICLv2 takes 24.5 H100-days, about 1,180 A100 hours \citep{tabiclv2}. TabICL takes 20 days on three A100s, 1,440 A100 hours \citep{qu2025tabicl}; TabPFNv2 takes about two weeks on eight RTX 2080 Ti, 2,690 GPU-hours \citep{hollmann2025tabpfnv2}, or about 460 A100 hours. TabPFN-3 does not report its pretraining compute \citep{tabpfn3}.

\paragraph{PFNs, per dataset.} In-context learning replaces fitting with one fit-and-predict pass. On BeyondArena-Lite, TabICLv2 takes 0.48 A100 hours per dataset on average (median 26 seconds). TabPFN-3 takes 0.70 GPU-hours on average (median 53 seconds).

\begin{table}[h]
\caption{Compute in A100 hours; CPU work converted at 61.5 vCPU-hours per A100 hour. Per-dataset costs are means over the 142 BeyondArena-Lite datasets, medians in parentheses.}
\label{tab:compute}
\begin{center}
\footnotesize
\begin{tabular}{lll}
\toprule
 & PriorTune & PFN \\
\midrule
Once per model & Grid: 400 (MLP) & Pretraining: 1,180 (TabICLv2) \\
 & \phantom{Grid: }300 (LightGBM-TabPrep) & \phantom{Pretraining: }1,440 (TabICL) \\
 & Selection: $<0.01$ (T), 10 (T+E, $K=200$) & \phantom{Pretraining: }460 (TabPFNv2) \\
 & & \phantom{Pretraining: }not reported (TabPFN-3) \\
\addlinespace
Per dataset & Fit $K=25$: 0.1 (0.04), LightGBM proxy & Fit and predict: 0.48 (0.007), TabICLv2 \\
\bottomrule
\end{tabular}
\end{center}
\end{table}

\subsubsection{Inference latency}
\label{app:latency}

\paragraph{Datasets and fits.} We measure latency (Figure~\ref{fig:pareto}) on 10 BeyondArena datasets that cover its dataset groups (Table~\ref{tab:latency-datasets}). For each dataset, we fix one outer train-test split. Under T, we deploy the validation-best configuration. Under T+E, we build the 40-step greedy ensemble of TabArena on the validation predictions. Ensemble members predict sequentially, and the ensemble averages their predictions. TabPack tensorizes the ensemble. In-context models use the inner training and validation splits as context and run at default settings, which include their internal ensembling: 8 estimators for TabPFN-3 and TabICLv2, and 16, 8, and 8 for the large, medium, and small Kumo-Tabular. The Elo of Figure~\ref{fig:pareto} comes from the full BeyondArena.

\begin{table}[h]
\caption{Datasets of the latency measurement. Rows are those of the outer training split.}
\label{tab:latency-datasets}
\begin{center}
\small
\begin{tabular}{lllrr}
    \toprule
    Dataset & Task & Split & Rows & Features \\
    \midrule
    credit-g & binary & i.i.d. & 666 & 20 \\
    mice-protein & multiclass & grouped & 720 & 77 \\
    coffee-rating & regression & temporal & 2{,}137 & 12 \\
    predict-students & multiclass & i.i.d. & 2{,}949 & 36 \\
    churn & binary & i.i.d. & 3{,}333 & 19 \\
    emscad & binary & i.i.d. & 11{,}640 & 17 \\
    video-transcoding & regression & grouped & 44{,}897 & 19 \\
    sdss17 & multiclass & i.i.d. & 52{,}035 & 11 \\
    sf-permit-time & regression & temporal & 60{,}784 & 37 \\
    kick & binary & temporal & 68{,}012 & 32 \\
    \bottomrule
\end{tabular}
\end{center}
\end{table}

\paragraph{Timing.} For each method and dataset, the fitted model is kept in memory. We do 5 warm-up predict calls on single test rows and then time 50 calls. We report the median over calls and then over the 10 datasets. All methods run on one host. GBDTs run on 16 CPU threads. DNNs and in-context models run on one GPU.

\paragraph{KV cache.} We time each PFN in three modes. In recompute, every call re-encodes the training context. With a key-value (KV) cache, the encoded context is computed once at fit time and kept on either the GPU or on the CPU. Table~\ref{tab:kv-cache} lists the options. The KV cache of TabPFN-3 is quantized to int8 by default, so its cached predictions differ from recompute by up to 0.14 in probability. For TabICLv2, they differ by up to 0.015.

\begin{table}[h]
\caption{Options of the in-context models for the three timed modes. TabPFN-3 and TabICLv2 take them in the constructor (\texttt{tabpfn} 9.0.0, \texttt{tabicl} 2.2.0). Kumo-Tabular takes them in the TabArena wrapper.}
\label{tab:kv-cache}
\begin{center}
\small
\setlength{\tabcolsep}{4pt}
\resizebox{\ifdim\width>\linewidth\linewidth\else\width\fi}{!}{%
\begin{tabular}{llll}
    \toprule
    Model & Recompute & KV cache on GPU & KV cache on CPU \\
    \midrule
    TabPFN-3 & \texttt{fit\_mode=} & \texttt{fit\_mode="fit\_with\_cache"} & \texttt{fit\_mode="fit\_with\_cache"} \\
     & \texttt{"fit\_preprocessors"}$^{a}$ & \texttt{keep\_cache\_on\_device=True} & \texttt{keep\_cache\_on\_device=False} \\
    TabICLv2 & \texttt{kv\_cache=False}$^{a}$ & \texttt{kv\_cache=True} & \texttt{kv\_cache=True}$^{b}$ \\
    Kumo-Tabular & \texttt{cache\_context=False} & \texttt{cache\_context=True}$^{c}$ & \texttt{cache\_context=True}$^{a}$ \\
    \bottomrule
\end{tabular}}
\end{center}
{\footnotesize $^{a}$Default. $^{b}$TabICL has no option to keep the cache on the CPU: we move it to the CPU after fit and copy it to the GPU in every call. $^{c}$The wrapper keeps the cache on the CPU: we move it to the GPU once after fit.\par}
\end{table}

\paragraph{Inference with TabPrep.} For TabPrep, we compile the feature generation in Numba and make it faster in two ways: (i) the fitted feature generators become lookup tables, and (ii) we compute only the generated features that the booster (LightGBM) splits on. The compiled predictor gives identical predictions. Without compilation, LightGBM-TabPrep under T costs 44 to 51\,ms per row through AutoGluon, against 0.8 to 0.9\,ms compiled, 51 to 54 times slower.

\subsection{Hyperparameter search spaces}
\label{app:comp-hps}

Every pool is a set of independent random draws from the search space below; no configuration is tuned on any dataset. The same draws are fitted on every synthetic and real task.

\paragraph{LightGBM.}
We use the LightGBM search space of TabArena \citep{erickson2025tabarena}, shown in Table~\ref{tab:lgbm_hpo_space}. Training follows the AutoGluon \texttt{LGBModel} semantics: up to 10,000 boosting rounds with AutoGluon's adaptive early stopping (offset 20, rate 0.3, maximum patience 300 rounds, scaled down above 10,000 training rows) on the validation split. The pool has $C=4{,}000$ configurations.

\begin{table}[h]
\caption{The hyperparameter tuning space for LightGBM.}
\label{tab:lgbm_hpo_space}
\begin{center}
\begin{tabular}{ll}
    \toprule
    Parameter & Distribution \\
    \midrule
    Learning rate & LogUniform[5e-3, 1e-1] \\
    Feature fraction & Uniform[0.4, 1.0] \\
    Bagging fraction & Uniform[0.7, 1.0] \\
    Bagging frequency & 1 \\
    \# leaves & LogUniformInt[2, 200] \\
    Min.\ data in leaf & LogUniformInt[1, 64] \\
    Extra trees & \{False, True\} \\
    \midrule
    Min.\ data per group & LogUniformInt[2, 100] \\
    Categorical L2 & LogUniform[5e-3, 2.0] \\
    Categorical smoothing & LogUniform[1e-3, 100] \\
    Max.\ categories to one-hot & LogUniformInt[8, 100] \\
    \midrule
    L1 regularization & Uniform[1e-4, 1.0] \\
    L2 regularization & Uniform[1e-4, 2.0] \\
    \midrule
    Max.\ boosting rounds & 10,000 \\
    Early stopping & AutoGluon adaptive (offset 20, rate 0.3, max.\ 300) \\
    \bottomrule
\end{tabular}
\end{center}
\end{table}

\paragraph{LightGBM with TabPrep.}
The same LightGBM space, with the TabPrep feature generators of \citet{tschalzev2026tabprep} prepended to the pipeline. Following the published protocol, TabPrep is enabled for one configuration in four; an enabled configuration also draws the eight TabPrep axes in Table~\ref{tab:tabprep_hpo_space}. The pool has $C=4{,}000$ configurations and shares its LightGBM draws with the plain LightGBM pool, so the two pools differ only in preprocessing.

\begin{table}[h]
\caption{The TabPrep axes drawn for the one-in-four configurations with TabPrep enabled.}
\label{tab:tabprep_hpo_space}
\begin{center}
\begin{tabular}{ll}
    \toprule
    Parameter & Distribution \\
    \midrule
    TabPrep enabled & Bernoulli(0.25) \\
    \midrule
    OpenFE max.\ features & \{2000, 1000\} \\
    OpenFE generation seed & \{42, 84, 168, 336, 672\} \\
    Cat.\ combination max.\ features & \{100, 500\} \\
    Cat.\ combination generation seed & \{42, 84, 168, 336, 672\} \\
    Random-subset feature combination \# subsets & \{50, 1, 100\} \\
    Random-subset generation seed & \{42, 84, 168, 336, 672\} \\
    OOF target encoding generation seed & \{42, 84, 168, 336, 672\} \\
    Group-by max.\ features & \{500, 100, 1000\} \\
    \bottomrule
\end{tabular}
\end{center}
\end{table}

\paragraph{MLP with cosine embeddings.}
The MLP is the independent member model of TabPack \citep{tabpack}: a feed-forward MLP with cosine numerical embeddings \citep{gorishniy2022embeddings}, trained with the Muon optimizer for the hidden weights and AdamW for the rest. The pool is a frozen registry of $C=4{,}000$ draws from the space in Table~\ref{tab:mlp_cos_hpo_space}. The seven sampled axes follow TabPack's cosine base model; the hidden width is fixed at 384. Members are trained individually with per-member optimizer steps, up to 256 epochs, and early stopping after 16 non-improving validations, without TabPack's online ensemble.

\begin{table}[h]
\caption{The hyperparameter tuning space for the MLP with cosine embeddings.}
\label{tab:mlp_cos_hpo_space}
\begin{center}
\begin{tabular}{ll}
    \toprule
    Parameter & Distribution \\
    \midrule
    \# blocks & UniformInt[1, 4] \\
    Block width & 384 \\
    Dropout rate & \{0.0, Uniform[0.0, 0.5]\} \\
    \midrule
    Embedding dim & UniformInt[8, 32], step=4 \\
    Embedding init.\ scale & LogUniform[1e-2, 1e1] \\
    \midrule
    Learning rate (AdamW) & LogUniform[1e-4, 5e-3] \\
    Learning rate (Muon) & LogUniform[1e-3, 1e-1] \\
    Weight decay & LogUniform[1e-3, 1.0] \\
    Batch size & dataset-dependent (TabArena rule) \\
    Max epochs & 256 \\
    Early-stopping patience & 16 \\
    \bottomrule
\end{tabular}
\end{center}
\end{table}

\paragraph{TabPack.}
TabPack \citep{tabpack} trains $64$ MLP members as one packed tensor program and combines them with the greedy online ensemble of the TabPack paper. The members are the first $64$ configurations of the MLP registry above. A meta-tuned TabPack is a pack of $K=64$ configurations selected from the same pool of 4,000 MLP configurations with the T+E objective and deployed with the same online ensemble; Figure~\ref{fig:pareto} and Figure~\ref{fig:main} use this subset.

\subsection{Synthetic priors}
\label{app:priors}

All three priors are structural causal models: a directed acyclic graph over latent variables, a random function per edge set, and an observation process that turns node values into a table. This section lists the three parts for each prior.

\subsubsection{Graph structure}

\paragraph{TabICLv2.} A random Cauchy DAG with $\mathrm{LogInt}(2,32)$ nodes. The probability of an edge from node $i$ to node $j$ is $\sigma(A+B_i+C_j)$, where $A$, $B_i$, $C_j$ are independent standard Cauchy variables: $A$ sets the overall density, $B_i$ and $C_j$ the out- and in-degree of each node. Features and the target are assigned to random nodes. A graph is resampled until the feature nodes and the target node share an ancestor.

\paragraph{TabICLv1.} A layered MLP graph as in TabPFNv1: $\mathrm{num\_layers}$ layers of width $\mathrm{hidden\_dim}$, both drawn from log-scaled truncated normals with per-dataset means in $[1,6]$ and $[5,130]$, with block-wise or Bernoulli dropout of edges. In causal mode (sampled per dataset) features and target are read from hidden activations; otherwise the root causes are the features and the last layer is the target.

\paragraph{TabPFNv1.} The same layered MLP graph, with layer count and width from gamma meta-distributions (means up to 6 and 300), $\mathrm{num\_causes}$ root variables, edge dropout, and the same causal or non-causal read-out. With probability $1/(1+e^{w})$, $w\sim U(2,10)$, the dataset is instead a draw from an RBF Gaussian process on Gaussian inputs \citep{hollmann2023tabpfn}.

\subsubsection{Edge functions}

\paragraph{TabICLv2.} Eight function types: MLP, tree ensemble (symmetric trees, $\mathrm{LogInt}(1,128)$ trees of depth 1 to 7), discretization to random centers, Gaussian process with a random kernel, linear, quadratic, EM-style cluster assignment, and product. The first three follow TabPFNv2, and the other five are new. With several parents, the prior either concatenates them and applies one function, or applies a function per parent and aggregates by sum, product, max, or log-sum-exp. Each node output is standardized, rescaled by a random feature-importance vector, and multiplied by a random node importance.

\paragraph{TabICLv1.} With probability 0.7 an MLP layer: a random affine map followed by an activation drawn from 19 fixed functions (Identity, Tanh, LeakyReLU, ELU, ReLU, ReLU6, SELU, SiLU, Softplus, Hardtanh, sign, sine, RBF, exponential, $\sqrt{|x|}$, $\1_{|x|\le1}$, $x^2$, $|x|$) or a random Fourier-feature function. Activations may differ between layers. With probability 0.3 a tree layer: an XGBoost multi-output regressor with at most 4 trees of depth at most 4, fitted on standard-normal targets and evaluated on the parent values.

\paragraph{TabPFNv1.} A random affine map followed by one activation per dataset, shared by all layers, drawn from Tanh, Identity, or ReLU.

\subsubsection{Transformations and filters}

\paragraph{TabICLv2.} Each scalar feature is min-max scaled and then warped by a Kumaraswamy function, with exponents drawn from $\mathrm{LogNum}(0.2,5)$. Categorical features come from discretizing sub-vectors of node values into 2 to 9 categories. Outliers are clamped at 4 standard deviations, and standardized values are clipped to $[-100,100]$. Constant columns are removed, and the feature order and class labels are permuted. Two filters reject datasets. The ancestor filter rejects a graph whose target shares no ancestor with the features. The ExtraTrees test fits 25 trees of depth 6 with bootstrapping and keeps a dataset only if their out-of-bag error beats the mean predictor in at least 95\% of 200 bootstrap resamples.

\paragraph{TabICLv1.} Gaussian noise is added after every layer, with a standard deviation drawn per dataset from $\mathrm{LogNum}(10^{-4},0.3)$. Root causes come from a normal, a uniform, or a mixed distribution. With probability 0.2, a dataset has categorical features. Each column then becomes categorical independently, with a uniformly drawn probability, and its values are binned by rank into $\max(2,\mathrm{round}(\Gamma(1,10)))$ categories. Outliers are clamped at 4 standard deviations, and the features and the regression target are standardized. Class labels are permuted. The original prior applies no filter. We apply the ExtraTrees test of TabICLv2, so that all three corpora are filtered alike.

\paragraph{TabPFNv1.} Gaussian noise is added at every node, with a mean and a standard deviation drawn per node. The standard deviation comes from $\mathrm{LogNum}(10^{-4},0.3)$. Categorical features are generated as in TabICLv1: with probability 0.2, binned by rank into $\Gamma(1,10)$ categories. Outliers are removed, and the features and the target are z-normalized. Features are rescaled by the fraction of columns in use. For classification, class boundaries are sampled from the data, and class labels are permuted. The original prior applies no filter. We apply the ExtraTrees test of TabICLv2.

\subsubsection{Sampling settings}

Table~\ref{tab:prior-settings} lists the sampling settings. TabICLv2 is the main prior of Section~\ref{subsec:prior}. TabICLv1 and TabPFNv1 share one shape specification: dataset $i$ has the same task type, row count, feature count, and class count in both sets, so a difference between them is a difference of generator alone.

\begin{table}[h]
\caption{Sampling settings of the three synthetic priors. LogU: log-uniform. Splits are train/validation/test and stratified for classification.}
\label{tab:prior-settings}
\begin{center}
\footnotesize
\begin{tabular}{llll}
\toprule
Setting & TabICLv2 & TabICLv1 & TabPFNv1 \\
\midrule
Generator & \texttt{graph\_scm} & \texttt{mix\_scm} & \texttt{prior\_bag} \\
Datasets & 4,000 & 2,000 & 2,000 \\
Task mix (reg/bin/multi) & 2,000 / 1,400 / 600 & 0.4 / 0.3 / 0.3 & 0.4 / 0.3 / 0.3 \\
Multiclass classes & $U\{3,\dots,10\}$ & $U\{3,\dots,10\}$ & $U\{3,\dots,10\}$ \\
Rows & LogU$[400, 60{,}000]$ & LogU$[1{,}024, 32{,}768]$ & LogU$[1{,}024, 32{,}768]$ \\
Features & $U[1, \mathrm{cap}(n)]$, cap $100\to20$ & LogU$[4, 64]$ & LogU$[4, 64]$ \\
Graph nodes & $\mathrm{LogInt}(2,32)$ & MLP, mean layers $[1,6]$ & MLP, gamma layers \\
Categorical share & $\mathrm{clip}(U(-0.5,1.2),0,1)$ & $p=0.2$ per dataset & $p=0.2$ per dataset \\
Node noise & none & $\mathrm{LogNum}(10^{-4},0.3)$ & $\mathrm{LogNum}(10^{-4},0.3)$ \\
Graph filter & ancestor test & none & none \\
Dataset filter & ExtraTrees, $p<0.05$ & ExtraTrees, $p<0.05$ & ExtraTrees, $p<0.05$ \\
Split & 64 / 16 / 20 & 64 / 16 / 20 & 64 / 16 / 20 \\
Attempts per dataset & $\le100$ & $\le64$ & $\le64$ \\
Base seed & 20260804 & 0 & 0 \\
\bottomrule
\end{tabular}
\end{center}
\end{table}

\FloatBarrier
\subsection{Additional results}
\label{app:ablation-grids}

\FloatBarrier
\subsubsection{Main results}
\label{app:main-improvement}

\begin{figure}[h]
    \centering
    \includegraphics[width=\linewidth]{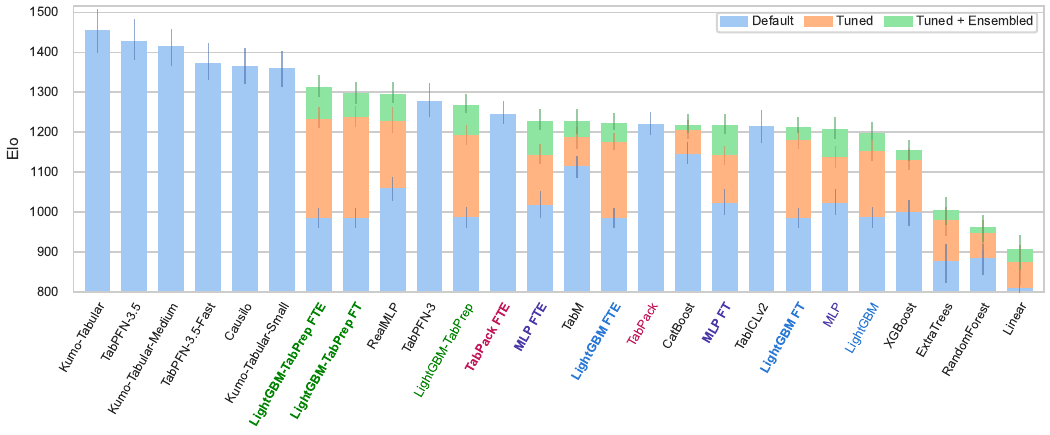}
    \caption{Elo of all 58 methods on the full BeyondArena under default, tuned, and tuned-and-ensembled deployment. Our families are coloured as in Figure~\ref{fig:pareto}.}
    \label{fig:elo-full}
\end{figure}

\paragraph{Full leaderboard.} Figure~\ref{fig:elo-full} shows the Elo of all 58 methods on the full BeyondArena.

\begin{figure}[h]
    \centering
    \includegraphics[width=\linewidth]{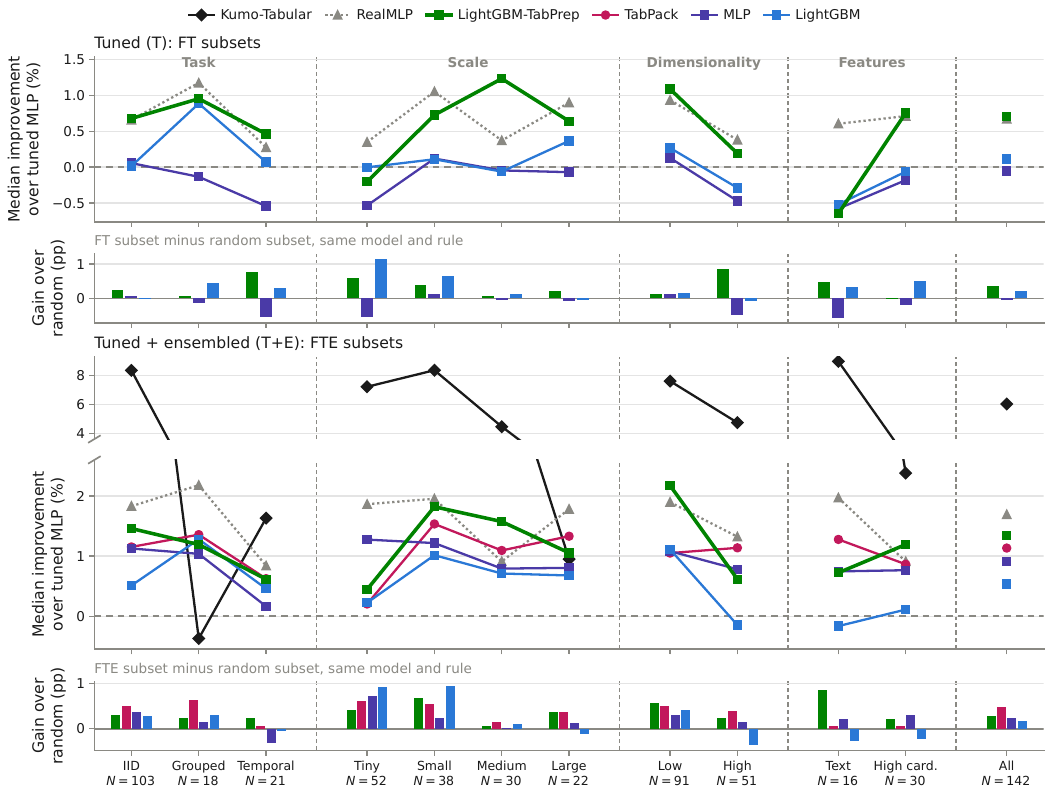}
    \caption{Median per-dataset improvement over the tuned MLP with a random subset on the full BeyondArena, by dataset group. Each dataset's improvement is averaged over its splits. \textbf{Top:} tuned (T), our models with their FT subsets. \textbf{Bottom:} tuned and ensembled (T+E), our models with their FTE subsets. Under each panel, the bars show the gain of each meta-tuned subset over the same model with its random subset, as the difference of the two medians.}
    \label{fig:main-improvement}
\end{figure}

\paragraph{Improvement over the MLP.} Figure~\ref{fig:main-improvement} repeats Figure~\ref{fig:main} with the improvement over the tuned MLP, $100\,(\ell_{\mathrm{base}}-\ell)/\ell_{\mathrm{base}}$. We average across the splits of each datasets and then take the median across datasets. Unlike Elo, the improvement keeps the size of each gain \citep{tschalzev2026lost} and does not depend on the set of compared methods. However, it depends on the choice of the baseline and averaging conceals skewness. Elo has the opposite properties, so neither metric suits every application \citep{tschalzev2026lost}. Under T, LightGBM-TabPrep with its FT subset has the highest median ($+0.71$). Under T+E, Kumo-Tabular ($+6.04$) and RealMLP ($+1.69$) lead LightGBM-TabPrep ($+1.34$). LightGBM-TabPrep leads Kumo-Tabular on grouped splits and on large datasets.

\FloatBarrier
\subsubsection{Selection rule}

Table~\ref{tab:selection} gives the win rate of every selection rule of Figure~\ref{fig:selection}.

\begin{table}[!htbp]
\caption{Win rate against random-$K$ of the selection rules of Section~\ref{subsec:selection-ablation}, synthetic out-of-fold (1,000 held-out tasks), T deployment, against random-$K$ baselines that no selection saw; mean over 200 repetitions ($\pm1.96$ standard errors of at most 0.0015). Draws $R$: our rule with $R$ validation resamples paired with $R$ references. Resampling kept: 16 validation resamples with one reference (Val.), 16 references on the original split (Ref.), or one of each (Neither). Bold: the best rule in the row.}
\label{tab:selection}
\begin{center}
\footnotesize
\setlength{\tabcolsep}{3.5pt}
\begin{tabular}{llcccccccccc}
\toprule
 & & \multicolumn{5}{c}{Draws $R$} & \multicolumn{3}{c}{Resampling kept} & \multicolumn{2}{c}{TabRepo} \\
\cmidrule(lr){3-7}\cmidrule(lr){8-10}\cmidrule(lr){11-12}
Model & $K$ & 16 & 8 & 4 & 2 & 1 & Val. & Ref. & Neither & Norm. & Count \\
\midrule
LightGBM & 25 & \textbf{0.636} & 0.634 & 0.632 & 0.627 & 0.620 & 0.630 & 0.630 & 0.621 & 0.617 & 0.621 \\
 & 64 & \textbf{0.621} & 0.618 & 0.613 & 0.607 & 0.599 & 0.614 & 0.609 & 0.600 & 0.588 & 0.602 \\
 & 200 & \textbf{0.599} & 0.596 & 0.592 & 0.585 & 0.576 & 0.593 & 0.586 & 0.574 & 0.561 & 0.575 \\
\addlinespace
LightGBM-TabPrep & 25 & \textbf{0.636} & \textbf{0.636} & 0.633 & 0.629 & 0.622 & 0.632 & 0.631 & 0.623 & 0.618 & 0.622 \\
 & 64 & \textbf{0.621} & 0.619 & 0.615 & 0.610 & 0.602 & 0.616 & 0.609 & 0.601 & 0.597 & 0.607 \\
 & 200 & \textbf{0.603} & 0.599 & 0.594 & 0.588 & 0.579 & 0.597 & 0.588 & 0.577 & 0.571 & 0.588 \\
\addlinespace
MLP & 25 & \textbf{0.697} & 0.696 & 0.692 & 0.687 & 0.679 & 0.691 & 0.689 & 0.681 & 0.687 & 0.693 \\
 & 64 & 0.669 & 0.665 & 0.661 & 0.654 & 0.646 & 0.664 & 0.655 & 0.645 & 0.658 & \textbf{0.672} \\
 & 200 & 0.631 & 0.626 & 0.619 & 0.612 & 0.601 & 0.626 & 0.613 & 0.600 & 0.616 & \textbf{0.633} \\
\bottomrule
\end{tabular}

\end{center}
\end{table}

\begin{figure}[!htbp]
    \centering
    \includegraphics[width=\linewidth]{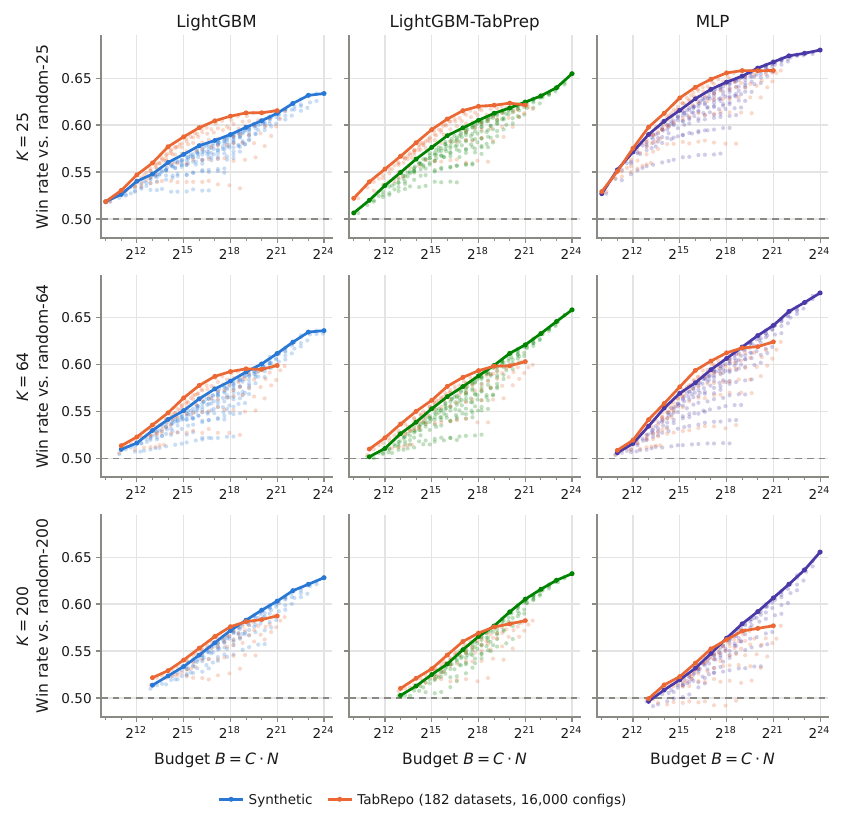}
    \caption{Synthetic versus TabRepo meta-training data at equal budget, as in Figure~\ref{fig:budget}, for $K=25$, $64$, and $200$.}
    \label{fig:budget-grid}
\end{figure}

\FloatBarrier
\subsubsection{Synthetic versus real meta-training data}

Figure~\ref{fig:budget-grid} repeats Figure~\ref{fig:budget} for $K\in\{25,64,200\}$. At $K=25$, TabRepo is ahead at equal budget up to $2^{20}$ to $2^{22}$ for all three models. The synthetic tasks end above its best for all three (Figure~\ref{fig:budget-grid}).

\paragraph{TabRepo overlap.} Of the 200 TabRepo datasets we exclude 18. Ten are content-identical to or listed as overlapping with a TabArena dataset: Amazon\_employee\_access, APSFailure, Bioresponse, blood-transfusion-service-center, diabetes, Diabetes130US, diamonds, hiva\_agnostic, jm1, splice. Eight share the source data of a TabArena dataset under a different curation: bank-marketing, churn, credit-g, houses, qsar-biodeg, wine-quality-red, wine-quality-white, wine\_quality.

\begin{figure}[!htbp]
    \centering
    \includegraphics[width=\linewidth]{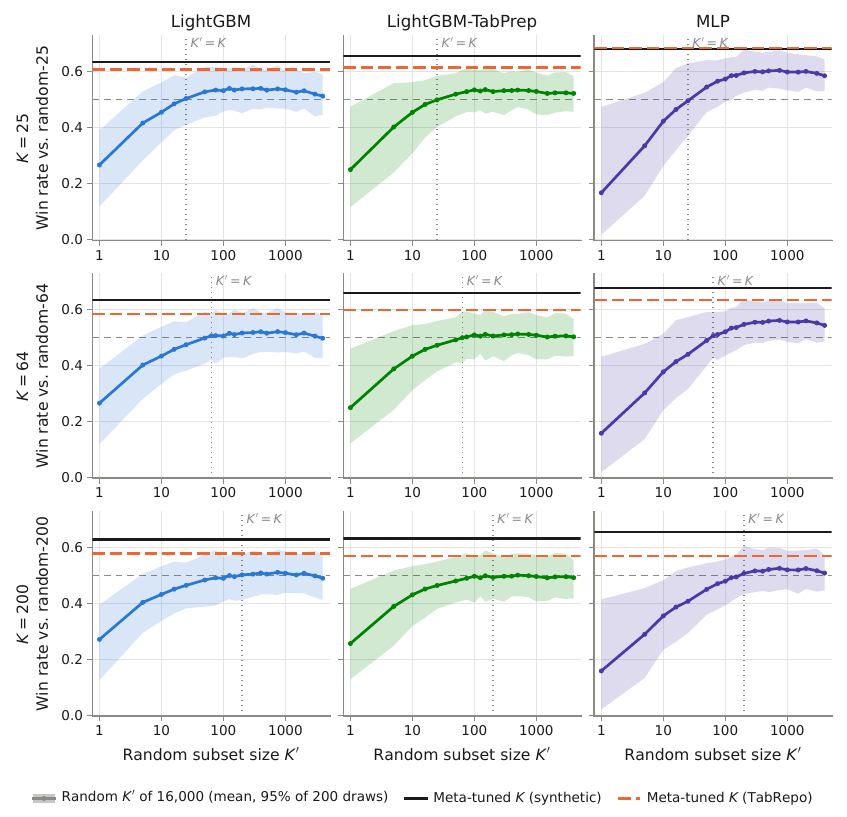}
    \caption{Random subsets of growing size $K'$ against the random-$K$ baseline, as in Figure~\ref{fig:ladder}, for $K=25$, $64$, and $200$.}
    \label{fig:ladder-grid}
\end{figure}

\FloatBarrier
\subsubsection{Meta-tuning versus more random search}

Figure~\ref{fig:ladder-grid} repeats Figure~\ref{fig:ladder} for $K\in\{25,64,200\}$.

\FloatBarrier
\subsubsection{Choice of prior}

\begin{figure}[!htbp]
    \centering
    \includegraphics[width=\linewidth]{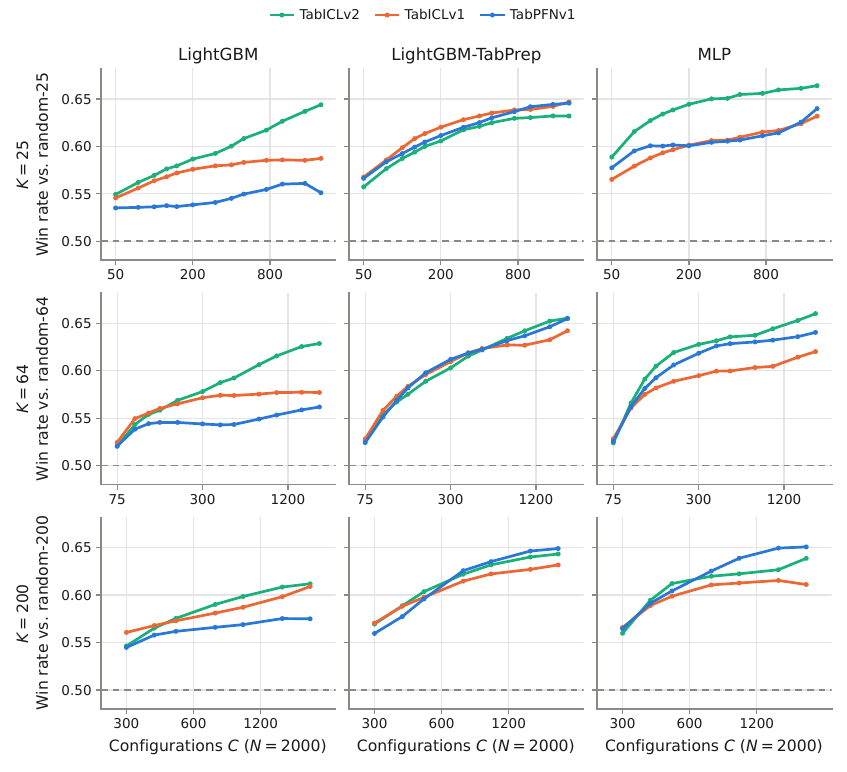}
    \caption{Three priors with 2,000 configurations on 2,000 synthetic tasks, as in the top row of Figure~\ref{fig:priors}: win rate against the number of configurations $C$ at $N=2{,}000$, for $K=25$, $64$, and $200$.}
    \label{fig:priors-grid}
\end{figure}

\begin{figure}[!htbp]
    \centering
    \includegraphics[width=\linewidth]{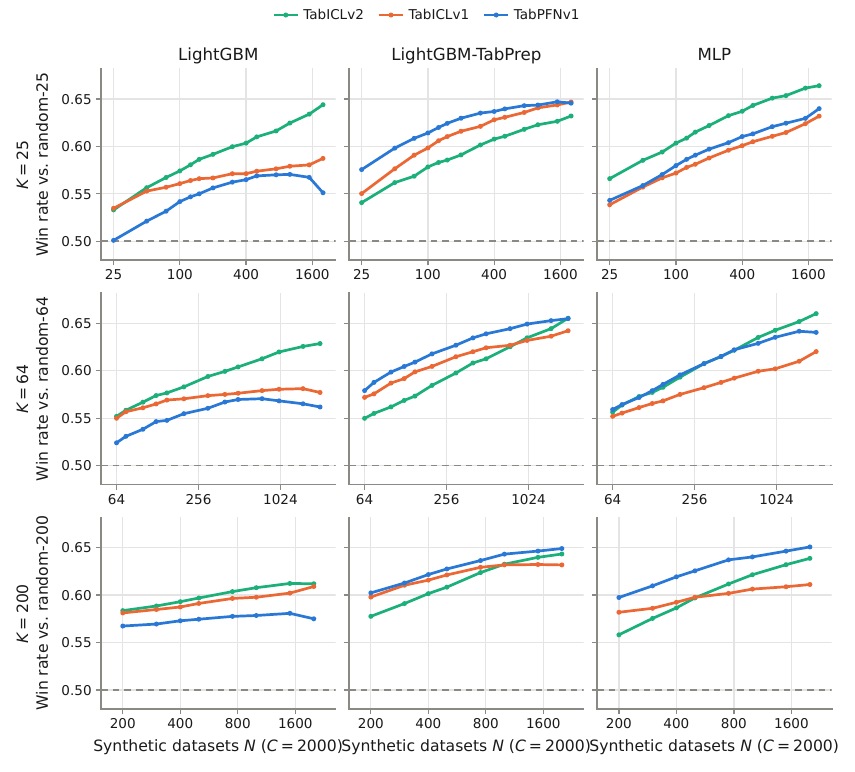}
    \caption{As Figure~\ref{fig:priors-grid}, against the number of synthetic tasks $N$ at $C=2{,}000$ (bottom row of Figure~\ref{fig:priors}).}
    \label{fig:priors-grid-datasets}
\end{figure}

Figures~\ref{fig:priors-grid} and~\ref{fig:priors-grid-datasets} repeat the top and bottom rows of Figure~\ref{fig:priors} for $K\in\{25,64,200\}$.

\FloatBarrier
\subsubsection{Scaling}

\begin{figure}[!htbp]
    \centering
    \includegraphics[width=\linewidth]{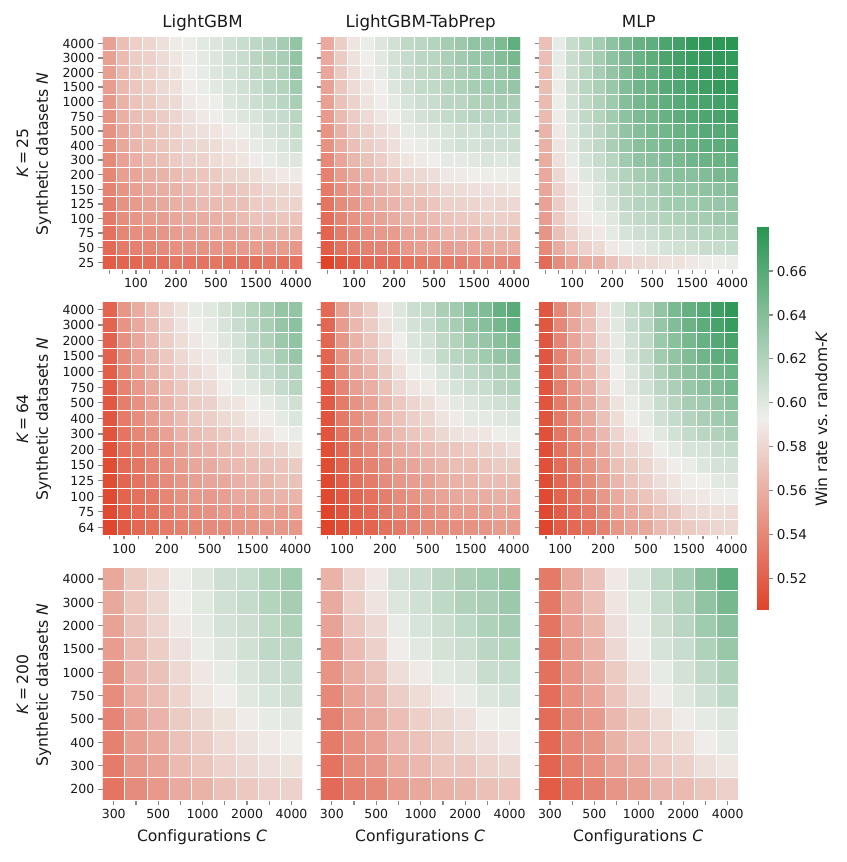}
    \caption{Win rate over the $(C,N)$ grid, as in Figure~\ref{fig:scaling}, for $K=25$, $64$, and $200$.}
    \label{fig:scaling-heatmaps-grid}
\end{figure}

\begin{figure}[!htbp]
    \centering
    \includegraphics[width=\linewidth]{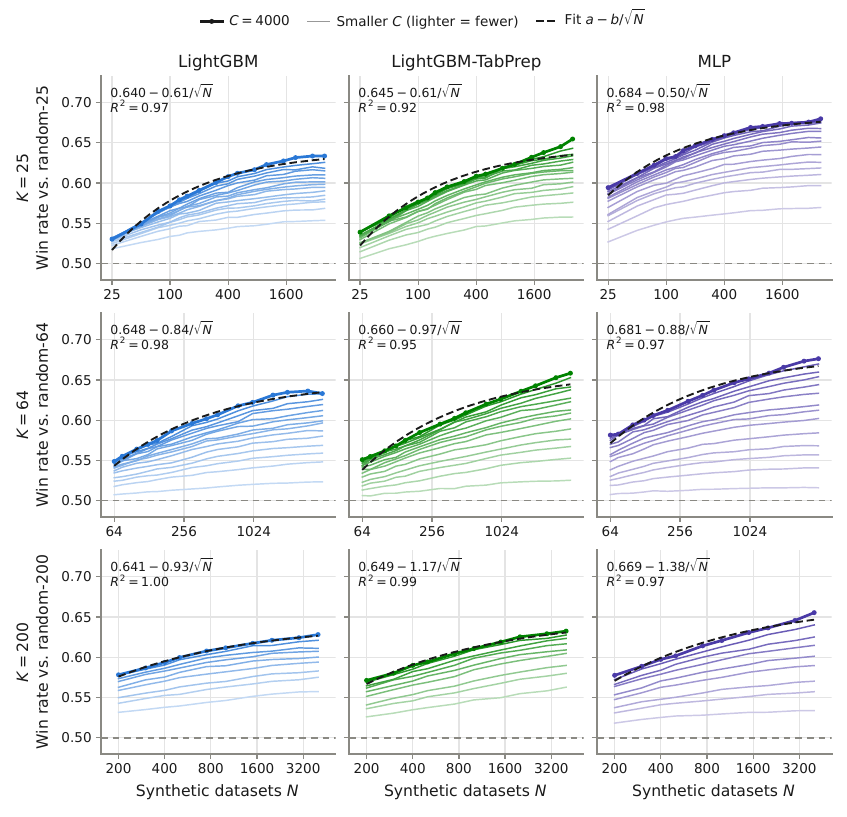}
    \caption{Win rate against the number of synthetic datasets $N$, one curve per pool size $C$, with the fit $a-b/\sqrt N$ at $C=4{,}000$, for $K=25$, $64$, and $200$.}
    \label{fig:scaling-grid-datasets}
\end{figure}

\begin{figure}[!htbp]
    \centering
    \includegraphics[width=\linewidth]{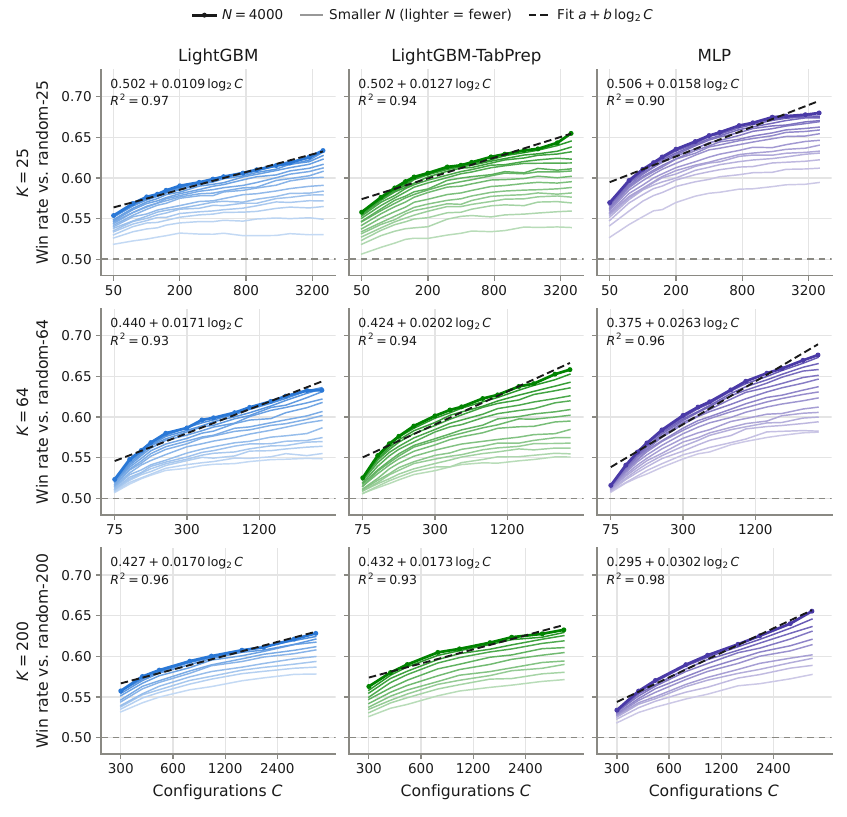}
    \caption{Win rate against the number of configurations $C$, one curve per number of synthetic datasets $N$, with the fit $a+b\log_2C$ at $N=4{,}000$, for $K=25$, $64$, and $200$.}
    \label{fig:scaling-grid-configs}
\end{figure}

Figure~\ref{fig:scaling-heatmaps-grid} shows the win rate over the $(C,N)$ grid, as in Figure~\ref{fig:scaling}, for $K\in\{25,64,200\}$. Figures~\ref{fig:scaling-grid-datasets} and~\ref{fig:scaling-grid-configs} show the curves along $N$ and along $C$.

\FloatBarrier
\subsubsection{Selected hyperparameters}

\begin{figure}[!htbp]
    \centering
    \includegraphics[width=\linewidth]{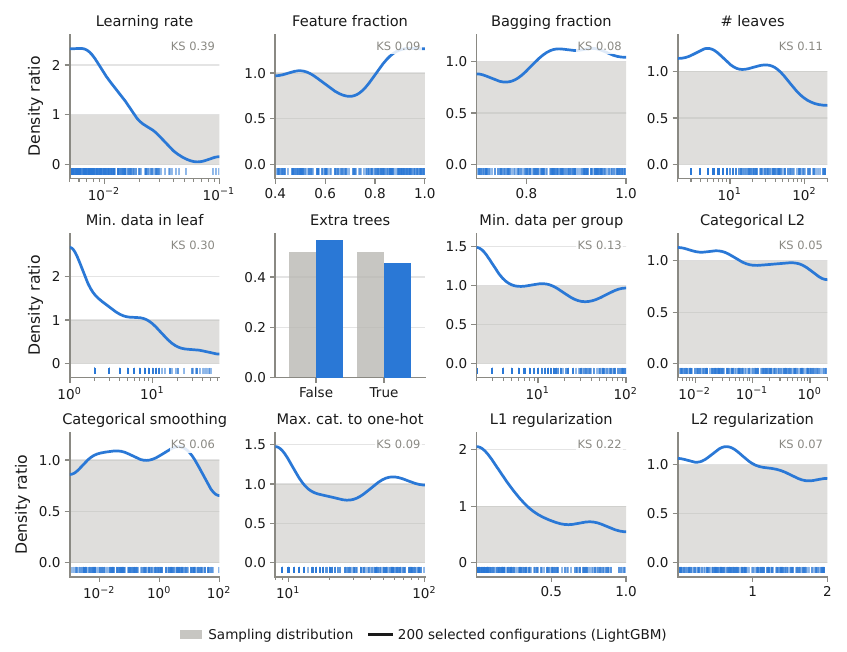}
    \caption{All hyperparameters of LightGBM: the 200 selected configurations against the sampling distribution, as in Figure~\ref{fig:hp}.}
    \label{fig:hp-lgbm-pure}
\end{figure}

\begin{figure}[!htbp]
    \centering
    \includegraphics[width=\linewidth]{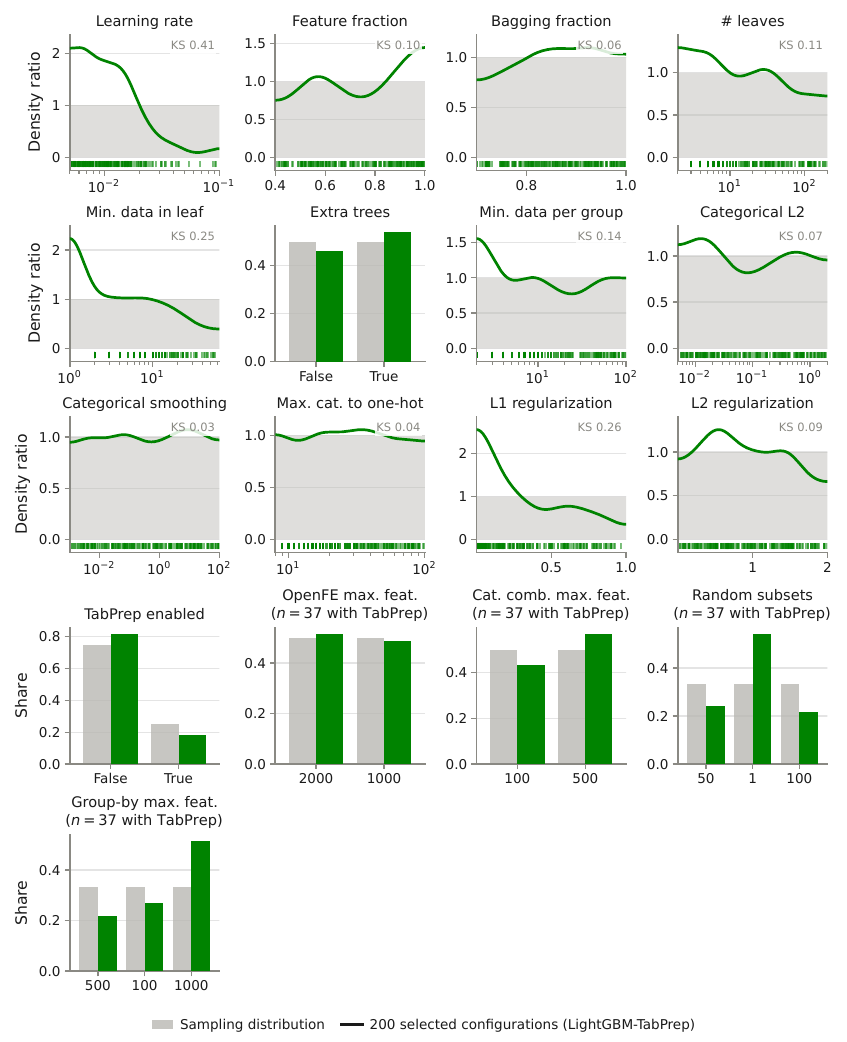}
    \caption{All hyperparameters of LightGBM-TabPrep, as in Figure~\ref{fig:hp}. The TabPrep settings count only the 37 selected configurations with TabPrep enabled.}
    \label{fig:hp-lgbm-prep}
\end{figure}

\begin{figure}[!htbp]
    \centering
    \includegraphics[width=\linewidth]{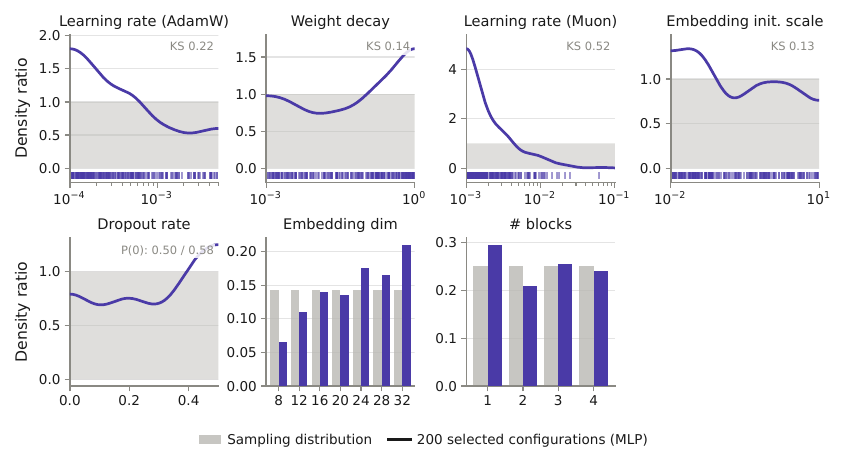}
    \caption{All hyperparameters of the MLP, as in Figure~\ref{fig:hp}. The dropout rate is 0 with probability 0.5 and uniform otherwise; the label gives the share at 0 (sampled / selected), and the curve is the continuous part.}
    \label{fig:hp-mlp}
\end{figure}

Figures~\ref{fig:hp-lgbm-pure}--\ref{fig:hp-mlp} show all hyperparameters of each model, as in Figure~\ref{fig:hp}.

\clearpage 
\section{Theory}
\label{app:theory}

\subsection{Setup and main propositions}
\label{app:theory-results}

\subsubsection{Setup}
\label{app:setup}

Fix a model $f$, a learning algorithm $\mathcal A$, a candidate pool $\mathcal C$ of $C$ hyperparameter configurations, and a deployment budget $1\le K\le C$. A \emph{subset} is $\mathcal S\in\Pi_K:=\{\mathcal S\subseteq\mathcal C:|\mathcal S|=K\}$. The goal is to choose one subset that performs well on the target distribution of real tasks, using synthetic tasks for meta-training. Performance is win rate against a uniformly random subset of the same size. All statements condition on the candidate pool.

\paragraph{Tasks and deployment.} Let $\mathcal P$ and $\mathcal Q$ be the target (e.g. TabArena or BeyondArena benchmarks \citep{erickson2025tabarena, beyondarena}) and source (e.g. sampled from the TabICLv2 prior \citep{tabiclv2}) distributions over tasks. Sample a target task $\tau\sim\mathcal P$, which specifies a row distribution $P_\tau$. Sample the rows of $D$ independently from $P_\tau$ and split the dataset into training, validation, and test sets of prescribed sizes. For each configuration, fit
\[
\hat f_c=\mathcal A(f,c,\xi,D),
\]
where $\xi$ collects the randomness in the dataset split and in the training of every configuration (initialization, subsampling, and any other stochastic step of $\mathcal A$). Each configuration is fitted once per task, and every subset that contains it, including the random reference it is compared against, reuses that fit. A \emph{deployment rule} $\mathcal E$ takes as an input a subset (its fitted members, train and validation subsets) and outputs a single predictor. We then score it on the test set:
\begin{equation}
L(\mathcal S;D,\hat f, \xi)
=\ell\!\left(\mathcal E(\mathcal S;D,\hat f,\xi);D_{\mathrm{test}}\right).
\label{eq:deployment}
\end{equation}
We require that $\mathcal E$ be deterministic given its inputs and that the same $\mathcal E$ be applied to a subset and to any subset it is compared against.

\paragraph{Two deployment rules.} We use the two regimes of \citet{erickson2025tabarena, beyondarena}. \emph{Tuned} (T) keeps the validation-best member,
\[
\mathcal E_{\mathrm T}(\mathcal S)=\hat f_{c^*},
\qquad
c^*=\argmin_{c\in\mathcal S}\ell(\hat f_c;D_{\mathrm{val}}).
\]
\emph{Tuned and ensembled} (T+E) runs $B$ steps of greedy ensemble selection with replacement on the validation split \citep{caruana2004ensemble}. Writing $w_c$ for the share of the $B$ slots taken by configuration $c$,
\[
\mathcal E_{\mathrm{T+E}}(\mathcal S)=\sum_{c\in\mathcal S}w_c\,\hat f_c,
\qquad
w_c\ge0,\quad\sum_{c\in\mathcal S}w_c=1.
\]
The tuned regime is the special case of tuned and ensembled that puts all weight on one member.

\paragraph{Population objective.} Let $h(a,b)=\1_{a<b}+\tfrac12\1_{a=b}$. The per-task utility of a subset is
\begin{equation}
u(\mathcal S,\tau)=\E_{D,\hat f,\xi\mid\tau,\ \mathcal R\sim\mathrm{Unif}(\Pi_K)}
\left[h\!\left(L(\mathcal S;D,\hat f,\xi),L(\mathcal R;D,\hat f,\xi)\right)\right],
\label{eq:task-utility}
\end{equation}
and the population objective is
\begin{equation}
U_{\mathcal P}(\mathcal S)=\E_{\tau\sim\mathcal P}[u(\mathcal S,\tau)],
\qquad
\mathcal S^*_{\mathcal P}\in\argmax_{\mathcal S\in\Pi_K}U_{\mathcal P}(\mathcal S).
\label{eq:target-objective}
\end{equation}
A uniformly random subset scores $\E_{\mathcal S\sim\mathrm{Unif}(\Pi_K)}[U_{\mathcal P}(\mathcal S)]=1/2$ by exchangeability (proof in \S\ref{app:proofs}), so $U_{\mathcal P}(\mathcal S)-1/2$ is the gain over random search at the same budget.

\paragraph{Evaluations and statistics.} Given a task $\tau$, let $\omega=(D,\hat f,\xi)$ be the \emph{evaluation}, with law $\mu_\tau$ on a space $\Omega$, and write $L(\mathcal S;\omega)$ for the deployed loss \eqref{eq:deployment}. The deployment rule is deterministic given its inputs, so once $\omega$ is fixed the deployed loss of every subset is fixed: $L(\mathcal S;\cdot)$ is a map on $\Omega$, which we assume measurable, as the expectation in \eqref{eq:task-utility} already requires. Write $\bar{\mathcal P}=\E_{\tau\sim\mathcal P}[\mu_\tau]$ and $\bar{\mathcal Q}=\E_{\tau\sim\mathcal Q}[\mu_\tau]$ for the marginal laws of $\omega$. A \emph{statistic} is a measurable map $\Phi:\Omega\to\mathcal X$, and $\mathcal P_\Phi=\bar{\mathcal P}\circ\Phi^{-1}$, $\mathcal Q_\Phi=\bar{\mathcal Q}\circ\Phi^{-1}$ are the laws of $\Phi(\omega)$. In this notation, \eqref{eq:task-utility}--\eqref{eq:target-objective} read
\[
U_{\mathcal P}(\mathcal S)=\E_{\tau\sim\mathcal P}\,\E_{\omega\sim\mu_\tau}\,\E_{\mathcal R}
\!\left[h\!\left(L(\mathcal S;\omega),L(\mathcal R;\omega)\right)\right],
\qquad \mathcal R\sim\mathrm{Unif}(\Pi_K)\text{ independent of }(\tau,\omega).
\]

\subsubsection{Upper bound on the improvement}
\label{app:oracle}

Suppose we have access to the target distribution $\mathcal P$ itself, so $U_{\mathcal P}$ is known on all of $\Pi_K$ and selection carries no estimation error. We ask how many of the subsets in $\Pi_K$ improve upon the best subset seen so far. Suppose the $M=\binom CK$ subsets are revealed in uniformly random order. A subset improves the oracle's choice precisely when its utility sets a new record.

\begin{proposition}[Oracle record count]
\label{prop:records}
If the oracle utilities are distinct, the $i$th revealed subset sets a record with probability $1/i$ over the uniformly random reveal order. Counting the initial subset as a record,
\[
\E[\text{number of records among the }M\text{ subsets}]
=\sum_{i=1}^{M}\frac1i=\log M+O(1).
\]
\end{proposition}
The count is a sum of independent Bernoulli variables with mean $\log M+O(1)$ \citep{renyi1962records}: improvements arrive at rate $1/i$ and thin out as the search proceeds. Since $\log M=K\log(C/K)+O(K)$, at fixed $K\ll C$ doubling the pool adds about $K\log2$ of them. Meta-tuning thus is the problem of recovering enough of the $\log M$ improvements, having access only to the $N$ source tasks.

\subsubsection{Transfer from synthetic to real tasks}
\label{app:transfer}

The target distribution is not available for meta-tuning: we observe only a benchmark of real tasks, which we reserve for evaluation. We therefore select the subset on a source distribution $\mathcal Q$ of synthetic tasks and transfer it. We sample $N$ independent source tasks $\tau_i\sim\mathcal Q$ and one evaluation $(D_i,\hat f_i,\xi_i)$ for each. Write $L_i(\mathcal S)=L(\mathcal S;D_i,\hat f_i,\xi_i)$. The empirical counterpart of \eqref{eq:target-objective} is
\begin{equation}
\widehat U_N(\mathcal S)=\frac1N\sum_{i=1}^N
\E_{\mathcal R\sim\mathrm{Unif}(\Pi_K)}
\left[h\!\left(L_i(\mathcal S),L_i(\mathcal R)\right)\right],
\label{eq:empirical-objective}
\end{equation}
and $\E[\widehat U_N(\mathcal S)]=U_{\mathcal Q}(\mathcal S)$ for every fixed subset. Its expectation is $U_{\mathcal Q}$, not $U_{\mathcal P}$; we quantify the price of this approximation.

\begin{assumption}[Comparison fidelity]
\label{as:transfer}
Source and target subset comparisons satisfy
\begin{equation}
\sup_{\mathcal S,\mathcal T\in\Pi_K}
\left|[U_{\mathcal P}(\mathcal S)-U_{\mathcal P}(\mathcal T)]
-[U_{\mathcal Q}(\mathcal S)-U_{\mathcal Q}(\mathcal T)]\right|
\le\varepsilon.
\label{eq:comparison-fidelity}
\end{equation}
\end{assumption}

\begin{proposition}[Transfer of subset selection]
\label{prop:transfer}
Let $\widehat{\mathcal S}$ be the subset returned by the selector and
\[
\eta_N=\max_{\mathcal S\in\Pi_K}\widehat U_N(\mathcal S)-\widehat U_N(\widehat{\mathcal S})
\]
its suboptimality on the source tasks. Under Assumption~\ref{as:transfer}, with probability at least $1-\delta$,
\begin{equation}
U_{\mathcal P}(\mathcal S^*_{\mathcal P})-U_{\mathcal P}(\widehat{\mathcal S})
\le\varepsilon+2\sqrt{\frac{\log(2\binom CK/\delta)}{2N}}+\eta_N.
\label{eq:transfer-bound}
\end{equation}
\end{proposition}

The bound separates selection error, source-target mismatch, and finite-task uncertainty. The number of source tasks reduces the second term, of order $\sqrt{\log M/N}$, but cannot remove systematic reversals of subset preferences ($\varepsilon$) or the error of the selection algorithm ($\eta_N$). We next investigate when Assumption~\ref{as:transfer} holds.

\begin{lemma}[Comparison discrepancy through a statistic]
\label{lem:tv}
The comparison \emph{factors through} $\Phi$ if for every $(\mathcal S,\mathcal R)\in\Pi_K\times\Pi_K$ there is a measurable $g_{\mathcal S,\mathcal R}:\mathcal X\to[0,1]$, depending on $(\mathcal S,\mathcal R)$ only, such that
\begin{equation}
h\!\left(L(\mathcal S;\omega),L(\mathcal R;\omega)\right)
=g_{\mathcal S,\mathcal R}\!\left(\Phi(\omega)\right)
\qquad\text{for every }\omega\in\Omega.
\label{eq:factor}
\end{equation}
If the comparison factors through $\Phi$, then Assumption~\ref{as:transfer} holds with $\varepsilon=2\operatorname{TV}(\mathcal P_\Phi,\mathcal Q_\Phi)$, where $\operatorname{TV}(P,Q)=\sup_A|P(A)-Q(A)|$. If moreover $\Phi=\kappa\circ\Psi$ for a statistic $\Psi:\Omega\to\mathcal Y$ and a measurable $\kappa:\mathcal Y\to\mathcal X$, then the comparison also factors through $\Psi$, and $\operatorname{TV}(\mathcal P_\Phi,\mathcal Q_\Phi)\le\operatorname{TV}(\mathcal P_\Psi,\mathcal Q_\Psi)$.
\end{lemma}

\begin{proposition}[Rank sufficiency]
\label{prop:statistic}
Let $\Phi(\omega)$ be the ordering of the $M$ losses $\{L(\mathcal S;\omega)\}_{\mathcal S\in\Pi_K}$, ties recorded. Then, for any deployment rule, Assumption~\ref{as:transfer} holds with $\varepsilon=2\operatorname{TV}(\mathcal P_\Phi,\mathcal Q_\Phi)$. Under $\mathcal E_{\mathrm T}$ the same holds for the finer statistic $\Psi(\omega)$, the validation and test orderings of the pool $\mathcal C$.
\end{proposition}

Hence what must transfer is the ranking of subsets by deployed loss. Under T this reduces to the validation and test ordering of individual configurations. Whether the synthetic prior preserves these orderings is an empirical question.

\subsubsection{Analysis of the forward-selection heuristic}
\label{app:heuristic}

The inner expectation in \eqref{eq:empirical-objective} runs over $\binom CK$ references, so in practice we draw $R$ independent references $\mathcal R_r\sim\mathrm{Unif}(\Pi_K)$ and optimize
\begin{equation}
\widetilde U_{N,R}(\mathcal S)=\frac1{NR}
\sum_{i=1}^N\sum_{r=1}^R
h\!\left(L_i(\mathcal S),L_i(\mathcal R_r)\right).
\label{eq:objective}
\end{equation}

\begin{proposition}[Reference averaging]
\label{prop:reference}
Write $\E_{\mathcal R}$ and $\operatorname{Var}_{\mathcal R}$ for the average over the $R$ reference draws with the source tasks held fixed. For every fixed subset,
\begin{equation}
\E_{\mathcal R}\!\left[\widetilde U_{N,R}(\mathcal S)\right]=\widehat U_N(\mathcal S),
\qquad
\operatorname{Var}_{\mathcal R}\!\left(\widetilde U_{N,R}(\mathcal S)\right)\le\frac1{4R}.
\label{eq:reference-moments}
\end{equation}
\end{proposition}
References thus add no bias, and their sampling noise vanishes as $1/R$.

\paragraph{Forward selection.} Exhaustive search over $\binom CK$ subsets is impractical. Starting from $\mathcal S_0=\emptyset$, we run
\[
\begin{aligned}
c_k&\in\argmax_{c\in\mathcal C\setminus\mathcal S_{k-1}}
\widetilde U_{N,R}(\mathcal S_{k-1}\cup\{c\}),\\
\mathcal S_k&=\mathcal S_{k-1}\cup\{c_k\},\qquad k=1,\ldots,K.
\end{aligned}
\]
We return exactly $K$ configurations, even if an addition lowers the objective. The objective is not monotone in general, so forward selection is a heuristic with no generic greedy-optimality guarantee. In Proposition~\ref{prop:transfer}, both the gap between $\widetilde U_{N,R}$ and $\widehat U_N$ and the greedy suboptimality enter through $\eta_N$.

\subsection{Proofs}
\label{app:proofs}

\paragraph{Proof of the random-$K$ baseline.} Let $\mathcal S,\mathcal R$ be i.i.d.\ uniform on $\Pi_K$, independent of $(\tau,\omega)$, and write $a=L(\mathcal S;\omega)$, $b=L(\mathcal R;\omega)$, so that $\E_{\mathcal S}[U_{\mathcal P}(\mathcal S)]=\E_{\tau,\omega}\,\E_{\mathcal S,\mathcal R}[h(a,b)]$. For fixed $(\tau,\omega)$, the pairs $(\mathcal S,\mathcal R)$ and $(\mathcal R,\mathcal S)$ have the same law, so $\E_{\mathcal S,\mathcal R}[h(a,b)]=\E_{\mathcal S,\mathcal R}[h(b,a)]$, and adding the two with $h(a,b)+h(b,a)=1$ gives $\E_{\mathcal S,\mathcal R}[h(a,b)]=\tfrac12$ for every $(\tau,\omega)$, hence $\E_{\mathcal S}[U_{\mathcal P}(\mathcal S)]=\tfrac12$. \qed

\paragraph{Proof of Proposition~\ref{prop:records}.} Write $M=\binom CK$ and let $U_i=U_{\mathcal P}(\mathcal S_i)$ be the utility of the $i$th revealed subset. Let $\rho_i$ be the rank of $U_i$, from largest to smallest, among $U_1,\ldots,U_i$, so that $I_i=\1_{\rho_i=1}$ indicates a record. We first show that $\rho_1,\ldots,\rho_M$ are independent with $\rho_i$ uniform on $\{1,\ldots,i\}$; this is the classical record theorem of \citet{renyi1962records}.

Let $\pi$ be the permutation that sorts $(U_1,\ldots,U_M)$. Because the reveal order is uniform and the utilities are distinct, every ordering is equally likely, so $\pi$ is uniform over the $M!$ permutations. The rank vector $(\rho_1,\ldots,\rho_M)$ is a function of $\pi$, and this function is a bijection onto $\{1\}\times\{1,2\}\times\cdots\times\{1,\ldots,M\}$: it is injective because $\pi$ is recovered by inserting $U_i$ at position $\rho_i$ among $U_1,\ldots,U_{i-1}$ for $i=1,\ldots,M$, and both sets have $M!$ elements. Hence $(\rho_1,\ldots,\rho_M)$ is uniform on that product set, and for each marginal we can calculate
\[
\Pr(\rho_i=k_i)
=\sum_{k_1=1}^{1}\sum_{k_2=1}^{2}\cdots\sum_{k_{i-1}=1}^{i-1}
\sum_{k_{i+1}=1}^{i+1}\cdots\sum_{k_M=1}^{M}\frac1{M!}
=\frac{M!}{i}\cdot\frac1{M!}=\frac1i,
\]
since the retained sums have $1\cdot2\cdots(i-1)\cdot(i+1)\cdots M=M!/i$ terms. For independence, we observe that $\Pr(\rho_1=k_1,\ldots,\rho_M=k_M)=1/M!=\Pr(\rho_1=k_1)\cdots\Pr(\rho_M=k_M)$.

Therefore we also have that $\Pr(I_i=1)=\Pr(\rho_i=1)=1/i$, and $I_1,\ldots,I_M$ are independent because each is a function of a different $\rho_i$. By linearity, $\E\sum_iI_i=\sum_i\Pr(I_i=1)=H_M$, and comparing the sum with $\int x^{-1}dx$ gives $\log(M+1)\le H_M\le1+\log M$. Finally $(C/K)^K\le M\le(C/K)^Ke^K$ gives $\log M=K\log(C/K)+O(K)$. \qed

\paragraph{Proof of Proposition~\ref{prop:transfer}.} Fix $\mathcal{S}\in\Pi_K$. By \eqref{eq:empirical-objective}, $\hat U_N(\mathcal{S})$ is the mean of $N$ i.i.d.\ terms in $[0,1]$ with mean $U_{\mathcal{Q}}(\mathcal{S})$. Hoeffding's inequality gives $\Pr(|\hat U_N(\mathcal{S})-U_{\mathcal{Q}}(\mathcal{S})|>t)\le 2\exp(-2Nt^2)$, and a union bound over the $\binom{C}{K}$ subsets gives
\begin{equation}
\Pr\Big(\sup_{\mathcal{S}\in\Pi_K}|\hat U_N(\mathcal{S})-U_{\mathcal{Q}}(\mathcal{S})|>t\Big)
\le 2\binom{C}{K}\exp(-2Nt^2).
\end{equation}
Setting the right-hand side to $\delta$ and solving for $t$ yields $t_N=\sqrt{\log(2\binom{C}{K}/\delta)/(2N)}$. On the complementary event, which has probability at least $1-\delta$, let $\mathcal{S}^*$ maximize $U_{\mathcal{P}}$. Then
\begin{align*}
U_{\mathcal{P}}(\mathcal{S}^*)-U_{\mathcal{P}}(\hat{\mathcal{S}})
&\le U_{\mathcal{Q}}(\mathcal{S}^*)-U_{\mathcal{Q}}(\hat{\mathcal{S}})+\varepsilon
&&\text{by \eqref{eq:comparison-fidelity}}\\
&\le \hat U_N(\mathcal{S}^*)-\hat U_N(\hat{\mathcal{S}})+\varepsilon+2t_N
&&\text{by uniform concentration}\\
&\le\eta_N+\varepsilon+2t_N
&&\text{by the definition of }\eta_N.
\end{align*}
The bound holds for any selector. Ours maximizes the Monte Carlo objective $\widetilde U_{N,R}$ rather than $\widehat U_N$, and both that gap and the greedy heuristic's are part of $\eta_N$. \qed

\paragraph{Proof of Lemma~\ref{lem:tv}.} Since $\mathcal R\sim\mathrm{Unif}(\Pi_K)$ is independent of $(\tau,\omega)$ and all terms are bounded, the objective \eqref{eq:target-objective} and the factoring \eqref{eq:factor} give
\begin{equation}
\begin{aligned}
U_{\mathcal P}(\mathcal S)
&=\E_{\tau\sim\mathcal P}\,\E_{\omega\sim\mu_\tau}\,\E_{\mathcal R}
\!\left[h\!\left(L(\mathcal S;\omega),L(\mathcal R;\omega)\right)\right]\\
&=\E_{\mathcal R}\,\E_{\tau\sim\mathcal P}\,\E_{\omega\sim\mu_\tau}
\!\left[g_{\mathcal S,\mathcal R}(\Phi(\omega))\right]
=\E_{\mathcal R}\,\E_{\mathcal P_\Phi}[g_{\mathcal S,\mathcal R}],
\end{aligned}
\label{eq:full-expectation}
\end{equation}
where the last equality uses that $\bar{\mathcal P}=\E_{\tau\sim\mathcal P}[\mu_\tau]$ is the marginal law of $\omega$ and $\mathcal P_\Phi$ is the law of $\Phi(\omega)$ under it; likewise $U_{\mathcal Q}(\mathcal S)=\E_{\mathcal R}\,\E_{\mathcal Q_\Phi}[g_{\mathcal S,\mathcal R}]$. For any measurable $g:\mathcal X\to[0,1]$, the layer-cake formula gives
$|\E_Pg-\E_Qg|=|\int_0^1(P(g>t)-Q(g>t))\,dt|\le\operatorname{TV}(P,Q)$. Applying it to each $g_{\mathcal S,\mathcal R}$ and averaging over $\mathcal R$ gives $|U_{\mathcal P}(\mathcal S)-U_{\mathcal Q}(\mathcal S)|\le\operatorname{TV}(\mathcal P_\Phi,\mathcal Q_\Phi)$ for every $\mathcal S$. Applying this to $\mathcal S$ and $\mathcal T$ and using the triangle inequality yields \eqref{eq:comparison-fidelity} with $\varepsilon=2\operatorname{TV}(\mathcal P_\Phi,\mathcal Q_\Phi)$.

If $\Phi=\kappa\circ\Psi$, then \eqref{eq:factor} holds with $\Psi$ and $g_{\mathcal S,\mathcal R}\circ\kappa$ in place of $\Phi$ and $g_{\mathcal S,\mathcal R}$. Since $\{\Phi\in A\}=\{\Psi\in\kappa^{-1}(A)\}$, $|\mathcal P_\Phi(A)-\mathcal Q_\Phi(A)|=|\mathcal P_\Psi(\kappa^{-1}(A))-\mathcal Q_\Psi(\kappa^{-1}(A))|\le\operatorname{TV}(\mathcal P_\Psi,\mathcal Q_\Psi)$ for every $A$. \qed

\paragraph{Proof of Proposition~\ref{prop:statistic}.} Let $\Phi(\omega)=(\mathrm{rank}_\omega(\mathcal S))_{\mathcal S\in\Pi_K}$ be the vector of ranks of the losses $\{L(\mathcal S;\omega)\}_{\mathcal S\in\Pi_K}$, equal losses receiving equal rank; it takes values in a finite set. Since $h(a,b)$ depends only on whether $a<b$, $a=b$, or $a>b$, and dense ranks preserve this, $h(L(\mathcal S;\omega),L(\mathcal R;\omega))=\1_{\mathrm{rank}_\omega(\mathcal S)<\mathrm{rank}_\omega(\mathcal R)}+\tfrac12\1_{\mathrm{rank}_\omega(\mathcal S)=\mathrm{rank}_\omega(\mathcal R)}$, which is \eqref{eq:factor}. Lemma~\ref{lem:tv} gives $\varepsilon=2\operatorname{TV}(\mathcal P_\Phi,\mathcal Q_\Phi)$.

Under $\mathcal E_{\mathrm T}$, let $\Psi(\omega)$ be the pair of dense-rank vectors of the pool $\mathcal C$ under the validation losses and under the test losses. The deployed member of $\mathcal S$ is its validation-rank minimizer, ties broken by a fixed rule, and the deployed losses of two subsets compare as the test ranks of their deployed members, so $\Phi=\kappa\circ\Psi$ for a fixed $\kappa$. The second part of Lemma~\ref{lem:tv} applies. \qed

\paragraph{Proof of Proposition~\ref{prop:reference}.} Fix $\mathcal S$, hold the source tasks fixed, and set $Z_r=\frac1N\sum_{i=1}^N h(L_i(\mathcal S),\allowbreak L_i(\mathcal R_r))$, so that $\widetilde U_{N,R}(\mathcal S)=R^{-1}\sum_{r}Z_r$. The $Z_r$ are i.i.d.\ functions of the independent draws $\mathcal R_r$, lie in $[0,1]$, and have mean $\widehat U_N(\mathcal S)$ by \eqref{eq:empirical-objective}. Hence $\E_{\mathcal R}[\widetilde U_{N,R}(\mathcal S)]=\widehat U_N(\mathcal S)$, and $\operatorname{Var}_{\mathcal R}(\widetilde U_{N,R}(\mathcal S))=R^{-1}\operatorname{Var}(Z_1)\le1/(4R)$, since a variable supported on $[0,1]$ has variance at most $1/4$. \qed

\end{document}